\documentclass[runningheads]{llncs}
\usepackage[utf8]{inputenc}
\usepackage{authblk}
\usepackage{xcolor}         
\usepackage{setspace}
\usepackage{soul}
\usepackage{amssymb} 
\usepackage[margin=1.25in]{geometry}
\usepackage{graphicx}
\graphicspath{ {./figures/} }
\usepackage{subcaption}
\usepackage{amsmath}
\usepackage{lineno}
\usepackage{xspace}
\usepackage{newtxmath}
\DeclareMathAlphabet{\mathpzc}{T1}{pzc}{m}{it}
\usepackage{booktabs}       
\usepackage{multirow}
\linenumbers

\newcommand{\hide}[1]{}
\usepackage{colortbl}

\definecolor{Gray}{gray}{0.9}
\definecolor{LightCyan}{rgb}{0.88,1,1}
\newcolumntype{a}{>{\columncolor{Gray}}c}
\newcolumntype{b}{>{\columncolor{white}}c}

\usepackage{amsfonts}
\usepackage{mathtools}
\usepackage{url}
\usepackage{enumitem}
\usepackage{authblk}
\usepackage{comment}
\usepackage{mdframed}
\usepackage{wrapfig}
\usepackage{ntheorem}

\newcommand{\xhdr}[1]{\vspace{0em}\noindent{{\bf #1.}}}
\newcommand{\ie}{\textit{i.e., \xspace}}
\newcommand{\eg}{\textit{e.g., \xspace}}
\newcommand{\std}[1]{\scriptsize{$\pm$#1}}
\usepackage{colortbl}
\definecolor{Gray}{gray}{0.9}
\definecolor{LightCyan}{rgb}{0.88,1,1}
\newcolumntype{a}{>{\columncolor{Gray}}c}
\newcolumntype{b}{>{\columncolor{white}}c}

\usepackage{amsfonts}
\usepackage{enumitem}
\usepackage{comment}
\usepackage{mdframed}
\usepackage{wrapfig}
\usepackage{ntheorem}
\usepackage{makecell}
\usepackage{tikz}
\newcommand*\circled[1]{\tikz[baseline=(char.base)]{
            \node[shape=circle,draw,inner sep=1pt] (char) {#1};}}

\theoremseparator{}

\theoremstyle{nonumberplain}

\newmdtheoremenv[%
  backgroundcolor=white,
  linecolor=blue!60!black,
  linewidth=2pt,
  topline=true,
  rightline=false,
  skipabove=10pt,
  skipbelow=10pt,
  leftline=false]{ouresponse}{}
  
\newmdtheoremenv[%
  backgroundcolor=gray!40,
  linecolor=blue!50!black,
  linewidth=2pt,
  topline=false,
  rightline=false,
  skipabove=10pt,
  skipbelow=10pt,
  leftline=false]{ourbox}{}

\theoremstyle{nonumberplain}
\usepackage{hyperref}

\usepackage[style=nejm, 
citestyle=numeric-comp,
sorting=none]{biblatex}
\title{In-Context Explainers: Harnessing LLMs for Explaining Black Box Models}

\titlerunning{In-Context Explainers: Harnessing LLMs for Explaining Black Box Models}

\author{Nicholas Kroeger\inst{1}* \and
Dan Ley\inst{2}* \and
Satyapriya Krishna\inst{2} \and
Chirag Agarwal\inst{2} \and
Himabindu Lakkaraju\inst{2}}

\institute{University of Florida \and
Harvard University\\
\vspace{2mm}
*Equal contribution. Corresponding author: Nick Kroeger (\href{mailto:nkroeger@ufl.edu}{\texttt{nkroeger@ufl.edu}})}
\authorrunning{Kroeger and Ley et al.}

\date{}

\newcommand{\LayerIntGrad}{Layer Integrated Gradients\xspace}
\newcommand{\LayerGradShap}{Layer Gradient SHAP\xspace}
\newcommand{\LayerDeepLift}{Layer Deep Lift\xspace}
\newcommand{\LayerGradXAct}{Layer Gradient x Activation\xspace}
\newcommand{\LayerAct}{Layer Activation\xspace}
\newcommand{\LayerCond}{Layer Conductance\xspace}

\newcommand{\LIME}{LIME\xspace}
\newcommand{\SHAP}{SHAP\xspace}

\newcommand{\Grads}{Vanilla Gradients\xspace}
\newcommand{\SmoothGrad}{SmoothGrad\xspace}
\newcommand{\IntGrad}{Integrated Gradients\xspace}
\newcommand{\GradtimesInput}{Gradient x Input\xspace}

\newcommand{\llms}{LLMs\xspace}

\newcommand{\pc}{\textsc{P-ICL}\xspace}
\newcommand{\pcg}{\textsc{PG-ICL}\xspace}
\newcommand{\vanillaExp}{Perturb ICL\xspace}
\newcommand{\classicICL}{Prediction-based ICL\xspace}
\newcommand{\instructionICL}{Perturb+Guide ICL\xspace}
\newcommand{\iclExp}{Explain ICL\xspace}
\newcommand{\EICL}{E-ICL\xspace}

\newcommand{\blood}{Blood\xspace}
\newcommand{\compas}{Recidivism\xspace}
\newcommand{\adult}{Adult\xspace}
\newcommand{\credit}{Credit\xspace}
\newcommand{\heloc}{HELOC\xspace}
\newcommand{\topk}{top-$k$\xspace}

\newcommand{\gptthree}{\textsc{Gpt-3.5}\xspace}
\newcommand{\gptfour}{\textsc{Gpt-4}\xspace}
\newcommand{\gptfournew}{\textsc{Gpt-4-0125-preview}\xspace}

\begin{document}
\nolinenumbers
\maketitle

\begin{abstract}
    Recent advancements in Large Language Models (LLMs) have demonstrated exceptional capabilities in complex tasks like machine translation, commonsense reasoning, and language understanding. One of the primary reasons for the adaptability of LLMs in such diverse tasks is their in-context learning (ICL) capability, which allows them to perform well on new tasks by simply using a few task samples in the prompt. Despite their effectiveness in enhancing the performance of LLMs on diverse language and tabular tasks, these methods have not been thoroughly explored for their potential to generate post hoc explanations. In this work, we carry out one of the first explorations to analyze the effectiveness of LLMs in explaining other complex predictive models using ICL. To this end, we propose a novel framework, \emph{In-Context Explainers}, comprising of three novel approaches that exploit the ICL capabilities of LLMs to explain the predictions made by other predictive models. We conduct extensive analysis with these approaches on real-world tabular and text datasets and demonstrate that LLMs are capable of explaining other predictive models similar to state-of-the-art post hoc explainers, opening up promising avenues for future research into LLM-based post hoc explanations of complex predictive models.

\end{abstract}

\section{Introduction}
\label{sec:intro}

\looseness=-1 Large Language Models (LLMs) have become ubiquitous across various industries and are being increasingly employed for a large range of applications, including language understanding~\cite{brown2020language}, genomics~\cite{zvyagin2023genslms}, tabular medical data records~\cite{lee2020biobert,alsentzer2019publicly}, and drug discovery~\cite{li2024cancergpt}. While LLMs attain high performance and generalization capabilities for numerous tasks~\cite{wei2022emergent}, they have also increased their parameter sizes and the computational costs for additional fine-tuning on new downstream tasks. To alleviate this, recent works have shown that LLMs can learn new tasks using in-context learning (ICL), which allows them to perform well on new tasks by simply using a few task samples in the prompt~\cite{liu2023pre}.

In-context learning allows language models to dynamically understand, adapt, and generate responses based on the immediate context provided in the prompt, eliminating the need for expensive retraining or fine-tuning. This capability of LLMs enables them to adapt to diverse applications. By generalizing from patterns within the input, ICL supports problem solving and complex reasoning, offering a powerful tool for LLMs~\cite{akyurek2022learning}. Despite its effectiveness in enhancing the performance of LLMs, the ability of in-context learning for data interpretation and analysis remains underexplored. In particular, there is very little work on systematically analyzing the potential of using in-context learning to explain the behavior of other complex predictive models. Thus, the potential of LLMs to serve as reliable explainers and enhance the understanding of predictive models remains an open question.
\begin{figure*}
    \centering
    \includegraphics[width=0.95\textwidth]{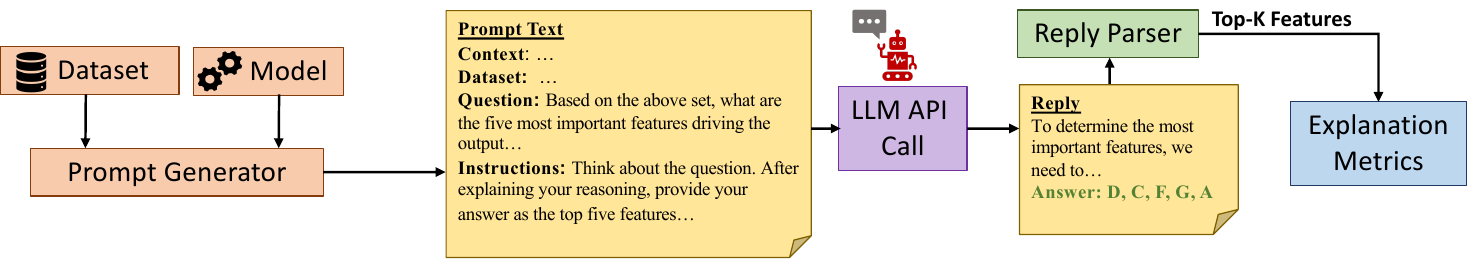}
    \vskip -0.1in
    \caption{\small
        \looseness=-1
        \xhdr{Overview of the in-context explanation generation and evaluation process} Given a dataset and a model to explain, we introduce novel ICL strategies to generate explanations of model predictions using LLMs. The resulting LLM-based explanations are then parsed, and their faithfulness is evaluated using diverse metrics.
    }
    \vspace{-0.15in}
    \label{fig:framework}
\end{figure*}

\looseness=-1\xhdr{Present work} In this work, we investigate whether in-context learning can help LLMs explain the behavior of other complex predictive models? (see Fig.~\ref{fig:framework}). To answer this question, we propose a novel framework called \emph{In-Context Explainers}. This framework comprises two broad in-context learning strategies: perturbation-based and explanation-based ICL, which provide the foundation for exploring the explanation-generating capabilities of state-of-the-art LLMs. Perturbation-based ICL builds on the ideas from explainable AI literature which involve constructing local linear approximations\hbox{~\cite{ribeiro2016should}} to explain the predictions made by a given ML model. To this end, we provide instance perturbations and their corresponding model outputs in the prompt to the LLM and ask the LLM to explain (i.e., output the top features driving) these predictions. We consider a couple of variants under this strategy, namely, \vanillaExp (\pc) and \instructionICL (\pcg). For \pc, we prompt the LLM to use its chain-of-thought reasoning. 
For \pcg, we add detailed guidelines in the prompt for generating explanations. In our \iclExp strategy, we prompt the LLM with a small, random selection of model input-output pairs and their respective explanations (generated using a state-of-the-art post hoc explanation method) to generate explanations for new samples.

\looseness=-1 We analyze the effectiveness of LLMs in explaining other predictive models on eight real-world datasets, four black box models, and three \textsc{Gpt} models. Our extensive investigation reveals the following key findings: \circled{\textbf{1}} ICL enables LLMs to generate faithful explanations that are on par with several state-of-the-art post hoc explanation methods (despite some of these methods having access to the underlying black box model); \circled{\textbf{2}} our proposed \iclExp prompting strategy allows LLMs to mimic the behavior of six state-of-the-art explanation methods; \circled{\textbf{3}} On average across five tabular datasets, we find that providing the LLM with detailed guidelines shows more consistent and higher faithfulness scores than simply providing generic instructions; \circled{\textbf{4}} Our results with three text datasets show that ICL capabilities aid LLMs to identify most important words in sentiment classification tasks.
Finally, our exploration highlights LLMs' effectiveness as post hoc explainers, paving the way for future research on LLM-generated explanations.
\section{Related Work}
\label{sec:relworks}
Our work lies at the intersection of post hoc explanations and LLMs, which we discuss below.

\looseness=-1
\xhdr{Post Hoc Explanations} The task of understanding model predictions has become increasingly intricate with the growing popularity of complex ML models~\cite{doshi2017towards} due to their inherent black box nature, which makes it difficult to interpret their internal reasoning. To this end, a plethora of feature attribution methods (commonly referred to as post hoc explanation methods) have been proposed to provide explanations for these models' predictions. These explanations are predominantly presented in the form of feature attributions, which highlight the importance of each input feature on the model's prediction. Broadly, post hoc explainers can be divided into perturbation-based and gradient-based methods. While perturbation-based methods~\cite{ribeiro2016should,lundberg2017unified,zeiler2014visualizing} leverage perturbations of the given instance to construct an interpretable approximation of the black box model behavior, gradient-based methods~\cite{smilkov2017smoothgrad,sundararajan2017axiomatic} leverage gradients \textit{w.r.t.} the given instance to explain model predictions. In this work, we primarily focus on state-of-the-art local post hoc explainers, \ie methods explaining individual feature importance for model predictions of individual instances.

\looseness=-1\xhdr{Large Language Models} LLMs have seen exponential growth in recent years, both in terms of their size and the complexity of tasks they can perform~\cite{radford17:learning}. Recent advances in LLMs are changing the paradigm of NLP research and have led to their widespread use across applications spanning machine translation~\cite{vaswani2017attention}, question-answering~\cite{brown2020language}, text generation~\cite{radford17:learning}, and medical data records~\cite{lee2020biobert,alsentzer2019publicly}. In this work, we, for the first time, explore their use in explaining other predictive models.

\section{Our Framework: In-Context Explainers}
\label{sec:method}
\looseness=-1
Here, we describe our proposed prompting strategies to generate natural language explanations from LLMs that can explain the behavior of predictive models. We will first discuss the notation used to describe the prompting templates and proceed to detail the prompts in Secs.~\ref{sec:vanilla}-\ref{sec:explanation}.

\looseness=-1
\xhdr{Notation} Let $f: \mathbb{R}^d \rightarrow [0,1]$ denote a black box ML model trained on a tabular data that takes an input $\mathbf{x} \in \mathbb{R}^d$ and returns the probability of $\mathbf{x}$ belonging to a class $c \in \mathcal{C}$ and the predicted label $y$. Following previous works in XAI~\cite{ribeiro2016should,smilkov2017smoothgrad}, we randomly sample points from the local neighborhood $\mathpzc{N}_{{x}}$ of the given input $\mathbf{x}$ to generate explanations, where $\mathpzc{N}_{{x}} = \mathcal{N}(0, \sigma^2)$ denotes the neighborhood of perturbations around $\mathbf{x}$ using a Normal distribution with mean 0 and variance $\sigma^2$.

\subsection{\vanillaExp}
\label{sec:vanilla}
Existing post hoc explainers, such as those relying on a large number of neighborhood samples~\cite{ribeiro2016should,smilkov2017smoothgrad}, often encounter computational bottlenecks. In contrast, we explore the utility of our proposed perturbation-based ICL strategy that leverages the power of LLMs in explaining the behavior of predictive models efficiently. In addition, unlike post hoc explainers like LIME~\cite{ribeiro2016should}, SmoothGrad~\cite{smilkov2017smoothgrad}, and SHAP~\cite{lundberg2017unified}, LLMs have the potential to produce natural language explanations that are more plausible and coherent to human practitioners. We aim to explore this by utilizing LLMs to articulate the top-$k$ most important features in determining the output of a given model $f$ in a rank-ordered manner. In particular, we sample input-output pairs from the neighborhood $\mathpzc{N}_{x}$ of $\mathbf{x}$ and generate their respective strings following a serialization template. For instance, let a neighborhood sample for $\mathbf{x}=[0.114, 0.111, 0.004, 0]$ be $\mathbf{x'}=[0.430, -0.001, -0.196, -0.139], \text{where}~\mathbf{x'}=\mathbf{x}+\delta~\text{and}~\delta=[0.316, -0.112, -0.200, -0.139]$, and assume the model predictions for $\mathbf{x}$ and $\mathbf{x'}$ are $1$ and $0$, respectively. Considering a binary classification problem, the corresponding change in the model predictions then belongs to $\{-1, 0, 1\}$. Next, we provide examples of transforming the perturbed sample ($\mathbf{x'}$) and the perturbation ($\delta$) into a natural-language string for our prompt templates. For brevity, in the following sections, we will use the perturbed sample in our template for describing our prompting strategies but explore both in our experiments.

\begin{figure}[t]
    \begin{ourbox}
        \footnotesize
        \textit{\\\textbf{{\# Using the Perturbed Sample $\mathbf{x'}$}}}\\
        Input: A: 0.430, B: -0.001, C: -0.196, D: -0.139, E: 0.050, F: 0.987\\
        Output: 0\vspace{-0.1in}
        
        \textit{\\\textbf{{\# Using the Raw Perturbation $\delta$}}}\\
        Change in Input: A: 0.316, B: -0.112, C: -0.200, D: -0.139, E: 0.050, F: -0.013\\
        Change in Output: 1
    \end{ourbox}
    \vspace{-0.1in}
    \caption{\small Sample serialization template for the Recidivism dataset with six features.}
    \label{fig:input_format}
\end{figure}
Motivated by the local neighborhood approximation works in XAI, the \vanillaExp prompting strategy presumes that the local behavior of model $f$ is a simple linear decision boundary, contrasting with the often globally exhibited complex non-linear decision boundary. Hence, assuming a sufficient number of perturbations in $\mathpzc{N}_{x}$, the LLM is expected to accurately approximate the black box model's behavior and utilize this information to identify the top-$k$ most important features. In addition, to alleviate the computational problems of post hoc explainers, we explore using a small number of $n_{\text{ICL}}$ samples (16 in our experiments) from $\mathpzc{N}_{x}$ in our prompting templates for LLMs to generate explanations. 
For samples in $\mathpzc{N}_{x}$, we select those with the highest predictive confidence by the underlying ML model, helping the LLM produce explanations centered on model certainty.

\looseness=-1 In our study, we explore two paradigms of generating the explanations: i) \textbf{\textit{\vanillaExp (\pc)}}, and ii) \textbf{\textit{\instructionICL (\pcg)}}, which we will describe next using the local neighborhood perturbations discussed above. Both of our strategies use four distinct steps in their template: \textit{Context}, \textit{Dataset}, \textit{Question}, and \textit{Instructions}. In \textit{Context}, we provide a general background of the underlying ML model, the number of features in the dataset, the number of classes, and the model predictions. 
In \textit{Dataset}, we leverage the success of ICL and provide a list of inputs and their respective model outputs using randomly sampled points from the local neighborhood of a given test input. In \textit{Question}, we specify the task we want the underlying LLM to perform. Finally, in the \textit{Instructions}, we enumerate the guidelines we want the LLM to follow while generating the output explanations. For \pc, we only prompt the LLM to leverage its chain-of-thought (CoT) reasoning abilities using ``\textit{Think about the question}'' before generating the response, but for \pcg, we prompt the model with a detailed guideline of how to think about the given problem.


\xhdr{\pc} Given the test sample $\mathbf{x}$ to be explained, we combine the \textit{Context} of the predictive model, the \textit{Dataset} of $n_{\text{ICL}}$ input-output pairs from $\mathpzc{N}_{x}$, the \textit{Question}, and the general \textit{Instructions} in our prompt to explore the effectiveness of the LLM in generating explanations. In Figure~\ref{fig:method1}, we provide a sample prompting template for the \pc strategy.
\begin{figure}[t]
    \begin{ourbox}
        \footnotesize
        \textit{\\\textbf{{\# \pc} prompt template}}\\
        \textbf{Context:} ``We have a two-class machine learning model that predicts based on 6 features: [`A', `B', `C', `D', `E', `F']. The dataset below contains the feature values `A' through `F' and the corresponding model outputs.''\\
        \textbf{Dataset:}\\
        Input: A: 0.192, B: 0.240, C: 0.118, D: 1.007, E: 0.091, F: 0.025\\
        Output: 0\\
        Input: A: 0.298, B: 0.256, C: 0.128, D: 1.091, E: -0.261, F: 0.168\\
        Output: 0\\
        \dots\\
        Input: A: 0.526, B: -0.298, C: -0.123, D: 1.078, E: -0.141, F: -0.110\\
        Output: 1\\
        \textbf{Question:} ``Based on the above set, what are the five most important features driving the output?''\\
        \textbf{Instructions:} ``Think about the question. After explaining your reasoning, provide your answer as the top five features ranked from most important to least important, in descending order, separated by commas. Only provide the feature names on the last line. Do not provide any further details on the last line.''\\\\
        \textbf{\# LLM Response:} To determine the most important features, we
need to \dots\\ \dots\\
        B, A, C, F, D, E
    \end{ourbox}
    \vspace{-0.1in}
    \caption{\small A sample prompt generated using our proposed \vanillaExp (\pc) prompting strategy.}
    \label{fig:method1}
\end{figure}

\looseness=-1\xhdr{\pcg} The \pcg prompting strategy transitions from specifying general instructions in the prompt to providing detailed guidance on the strategy for task execution. 
Rather than solely instructing the LLM to ``\textit{Think about the question}'' of what the task entails, this strategy delineates how to conduct the given task. The objective remains to explore the effectiveness of LLMs in explaining the behavior of other predictive models by identifying the top-$k$ most important features. 
However, with step-by-step guidelines, we aim to induce a more structured and consistent analytical process within the LLM to generate more faithful explanations. 
We follow a similar prompting template as in Method 1, including the four components, \textit{viz.,} \textit{Context}, \textit{Dataset}, \textit{Question}, and \textit{Instructions}, but we modify the \textit{Instructions} component. 
The \pcg prompt template is provided in Figure~\ref{fig:method2}.

\begin{figure}[h]
    \begin{ourbox}
        \footnotesize
        \textit{\\\textbf{{\# \pcg} prompt template}}\\
        \textbf{Context:} ``We have \dots outputs.''\\
        \textbf{Dataset:}\\
        Input: \dots \\
        \dots \\
        \textbf{Question:} ``Based on \dots output?''\\
        \textbf{Instructions:} ``For each feature, starting with `A' and continuing to `F':
        
        1. Analyze the feature in question. Rate the importance of the feature in determining the output on a scale of 0-100, considering both positive and negative correlations. Ensure to give equal emphasis to both positive and negative correlations and avoid focusing only on absolute values.
        
        2. After analyzing the feature, position it in a running rank compared to the features already analyzed. For instance, after analyzing feature `B', determine its relative importance compared to `A' and position it accordingly in the rank (e.g., BA or AB). Continue this process until all features from `A' to `F' are ranked.
        
        After explaining your reasoning, provide your answer as the final rank of features from `A' to `F' from most important to least important, in descending order, separated by commas. Only provide the feature names on the last line. Do not provide any further details on the last line.''\\\\
        \textbf{\# LLM Response:} To determine the most important features, we
need to \dots\\ \dots\\
        B, A, C, F, D, E
    \end{ourbox}
    \vspace{-0.1in}
    \caption{\small\looseness=-1 A sample prompt generated using the \instructionICL (\pcg) prompting strategy. Note that the \textit{Context}, \textit{Dataset}, and \textit{Question} are the same as in Fig.~\ref{fig:method1}.}
    \label{fig:method2}
\end{figure}

\looseness=-1 Here, we provide some detailed guidelines to the LLM for understanding the notion of important features and how to analyze them through the lens of correlation analysis. To achieve this, we instruct LLMs to study each feature sequentially and ensure that positive and negative correlations are equally emphasized. The LLM assigns an importance score to each feature in the given dataset and then positions it in a running rank. This rank encourages the LLM to differentiate features and avoid ties in its evaluation of feature importance. The final line in the template ensures that the LLM's responses are strictly analytical, minimizing non-responsiveness or digressions.

\subsection{\iclExp}
\label{sec:explanation}

\looseness=-1 Recent studies show that \llms can learn new tasks through ICL, enabling them to excel in new downstream tasks by merely observing a few instances of the task in the prompt. 
Unlike \vanillaExp (Sec.~\ref{sec:vanilla}), which samples input-output pairs from the neighborhood $\mathpzc{N}_{x}$ of a test sample $\mathbf{x}$ for ICL prompts, the \iclExp (\EICL) strategy uses random input-output-explanation triplets generated from a held-out dataset for prompting. 
Here, the explanations used in the ICL are generated by any post hoc explanation method as a ground truth. 
Intuitively, we explore whether an LLM can learn the behavior of a given post hoc explanation method by looking at some of its generated explanations. 

\looseness=-1 
For constructing the ICL set, we randomly select $n_{\text{ICL}}$ input instances $\mathbf{X}_{\text{ICL}}$ from the ICL split of the dataset and generate their predicted labels $\mathbf{y}_{\text{ICL}}$ using the given predictive model $f$. Next, we generate explanations $\mathbf{E}_{\text{ICL}}$ for samples ($\mathbf{X}_{\text{ICL}}$, $\mathbf{y}_{\text{ICL}}$) using any post hoc explainer. Using the above input-output-explanation triplets, we construct a prompt by concatenating them to form the prompt in Figure ~\ref{fig:method3}.
\begin{figure}[t]
    \begin{ourbox}
        \footnotesize\textit{\\\#~\textbf{\iclExp Prompt Template}}\\
        Input: A: 0.172, B: 0.000, C: 0.000, D: 1.000, E: 0.000, F: 0.000 \\
        Output: 1\\
        Explanation: A,C,B,F,D,E\\
        \dots\\
        Input: A: 0.052, B: 0.053, C: 0.073, D: 0.000, E: 0.000, F: 1.000 \\
        Output: 0\\
        Explanation: A,B,C,E,F,D\\
        Input: A: 0.180, B: 0.222, C: 0.002, D: 0.000, E: 0.000, F: 1.000 \\
        Output: 0\\
        Explanation:
    \end{ourbox}
    \vspace{-0.1in}
    \caption{\small An example prompt generated using the proposed \iclExp (\EICL) prompting strategy.}
    \label{fig:method3}
\end{figure}
The \EICL prompting strategy explores how LLMs can produce faithful explanations by analyzing the $n_{\text{ICL}}$ input-output-explanations generated by state-of-the-art post hoc explainer.
\section{Experimental Evaluation}
\label{sec:experiments}
Next, we evaluate the effectiveness of LLMs as post hoc explainers, focusing on three key questions: Q1) Can LLMs generate post hoc explanations for predictive models trained on tabular datasets? Q2) How well can LLMs generate explanations for sentiment classifiers? Q3) What impact do variations in the LLM's prompting strategy have on the faithfulness of explanations?

\subsection{Datasets and Experimental Setup}
\label{subsec:data_exps}
We first describe the datasets and models used to study the reliability of \llms as post hoc explainers and then outline the experimental setup.

\xhdr{Datasets} For tabular datasets, we follow previous LLM works~\cite{hegselmann2023tabllm} and perform analysis on five real-world tabular datasets: \textbf{\blood}~\cite{yeh2009knowledge}, \textbf{\compas}~\cite{compas}, \textbf{\adult}~\cite{adult}, \textbf{\credit}~\cite{default}, and \textbf{\heloc}~\cite{heloc}. For text datasets, we use three sentiment datasets: \textbf{Amazon reviews}, \textbf{IMDb}, and \textbf{Yelp}~\cite{misc_sentiment_labelled_sentences_331}. The datasets come with a random train-test split, and we further divide the train set, allocating 80\% for training and the remaining 20\% for ICL sample selection, as detailed in Sec.~\ref{sec:explanation}. See Appendix~\ref{app:details} for more details.

\looseness=-1
\xhdr{Predictive Models} For the tabular dataset classifiers, we consider three ML models with varying non-linearity in our experiments: Logistic Regression (LR), a three-layer Artificial Neural Network (ANN), and a six-layer ANN. For the sentiment classifier, we train a single layer transformer model for the text datasets to show the generalization capability of our analysis in generating explanations for diverse predictive models. We use \verb+PyTorch+~\cite{pytorch} to implement the LR, ANN, and transformer model. Please refer to Appendix~\ref{app:details} for architecture details and Tables~\ref{tab:performance} and~\ref{tab:performance_transformer} for predictive performances of these models.

\xhdr{Large Language Models} We consider \gptthree, \gptfour, and \gptfournew as LLMs for all experiments.
All experiments default to using \gptfour, unless stated otherwise.

\xhdr{Baseline Methods} For the tabular dataset classifiers, we use six post hoc explainers as baselines to investigate the effectiveness of explanations generated using \llms: \LIME~\cite{ribeiro2016should}, \SHAP~\cite{lundberg2017unified}, \Grads~\cite{zeiler2014visualizing}, \SmoothGrad~\cite{smilkov2017smoothgrad}, \IntGrad~\cite{sundararajan2017axiomatic}, and \GradtimesInput (ITG)~\cite{shrikumar2017learning}. 

For text classifiers, we cannot directly retrieve \textit{gradient-}based attributions at the input layer since the model uses an embedding layer with discrete vocabulary inputs. Hence, we compute attributions at the embedding layer of each token, where the core idea here is to use a layer-wise attribution method to compute how much each dimension of the embedding vector contributes to the final prediction. Finally, we sum and normalize attributions across all dimensions of each token's embedding vector, resulting in a single scalar value per token. The six gradient-based post hoc explainers for the text classifiers are  \LayerIntGrad (LIG), \LayerGradShap (LGS), \LayerDeepLift (LDL), \LayerGradXAct (LGxA), \LayerAct (LA), and \LayerCond (LC)~\cite{dhamdhere_how_2018} from the Captum python library \cite{kokhlikyan2020captum}.

\xhdr{Performance Metrics} We employ four distinct metrics to measure the faithfulness of an explanation. We use the Feature Agreement (FA) and Rank Agreement (RA) metrics introduced in~\cite{krishna2022disagreement} that compares the LLM's \topk directly with the LR model's coefficient (see Appendix~\ref{app:details} for details). The FA and RA metrics range from $[0,1]$, where 0 means no agreement and 1 means full agreement. In the absence of a \topk model coefficient (as is the case with ANNs), we use the Prediction Gap on Important feature perturbation (PGI) and the Prediction Gap on Unimportant feature perturbation (PGU) metrics from OpenXAI~\cite{agarwal2022openxai}. While PGI measures the change in prediction probability that results from perturbing the features deemed as influential, PGU examines the impact of perturbing unimportant features. 
Here, the perturbations are generated using Gaussian noise $\mathcal{N}(0, \sigma^2)$.
For text datasets, we adapt the PGI and PGU metrics to evaluate the faithfulness of explanations by measuring the impact of perturbing the \topk important and unimportant words respectively, as identified by the explanation model. The perturbations are implemented by masking the specified words rather than using Gaussian noise, which is more suited for continuous data-types. These modified metrics for text provide insights into how specific words influence the model's predictions. All reported values use the area under the curve (AUC) of the faithfulness score evaluated from $k=1$ to $k=3$, as a default setting, unless otherwise stated.

\xhdr{Implementation Details} To generate perturbations for each tabular dataset's ICL prompt, we use a neighborhood size of $\sigma{=}0.1$ and generate local perturbation neighborhoods $\mathpzc{N}_{{x}}$ for each test sample $\mathbf{x}$. 
We sample $\mathpzc{n}_{x}{=}10,000$ points for each neighborhood, where the values for $\sigma$ and $\mathpzc{n}_{x}$ were chosen to give an equal number of samples for each class, whenever possible.
For the text dataset's ICL prompt, we generate neighborhood sentences by randomly selecting a subset of words to omit, aiming to simulate a range of linguistic variations that enable sensitivity analysis of word-level changes in input sentences.
As shown in Figure ~\ref{fig:input_format}, we present the ICL set in two main formats: 1) as the perturbed sample ($\mathbf{x'}$) and its corresponding predicted output, or 2) as the perturbation ($\delta$) around a sample 
($\mathbf{x}$) and the corresponding change in output. 
Note that the perturbation ($\delta$) format for the ICL set is used as a default setting in all experiments unless otherwise stated.
Additionally, both of these formats are absent from Sec.~\ref{sec:explanation}, which uses test samples directly and does not compute perturbations.

\looseness=-1 For the LLMs, we use OpenAI's text generation API with a temperature of $\tau=0$ for our main experiments. To evaluate the LLM explanations, we extract and process its answers to identify the top-$k$ most important features. We first save each LLM query's reply to a text file and use a script to extract the listed features. We added explicit instructions like ``\textit{\dots provide your answer as a feature name on the last line. Do not provide any further details on the last line.}'' to ensure reliable parsing of LLM outputs. In rare cases, the LLM won't follow our requested response format or it replies with ``\textit{I don't have enough information to determine the most important features.}'' See Appendix~\ref{app:details} for further details. 

\begin{figure*}[h]
    \centering
    \textsc{\footnotesize Logistic Regression}
    \includegraphics[width=0.99\linewidth]{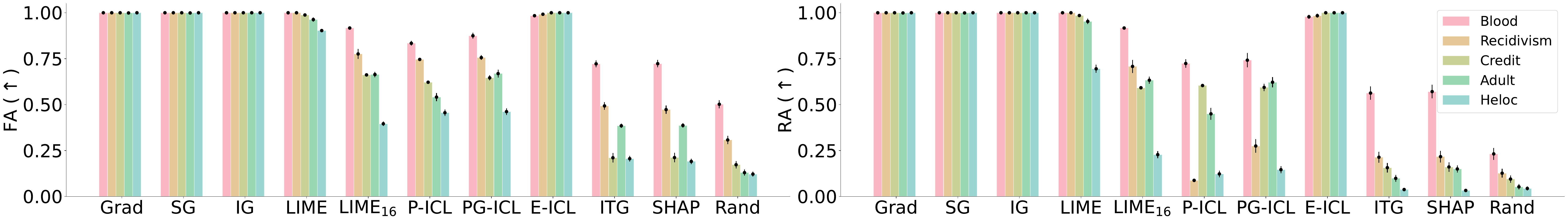}

    \textsc{\footnotesize Three-Layer ANN}
    \includegraphics[width=0.99\linewidth]{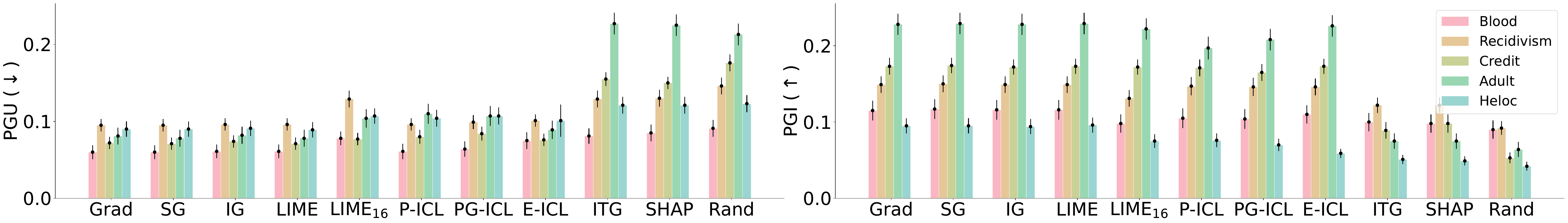}
    
    \textsc{\footnotesize Six-Layer ANN}
    \includegraphics[width=0.99\linewidth]{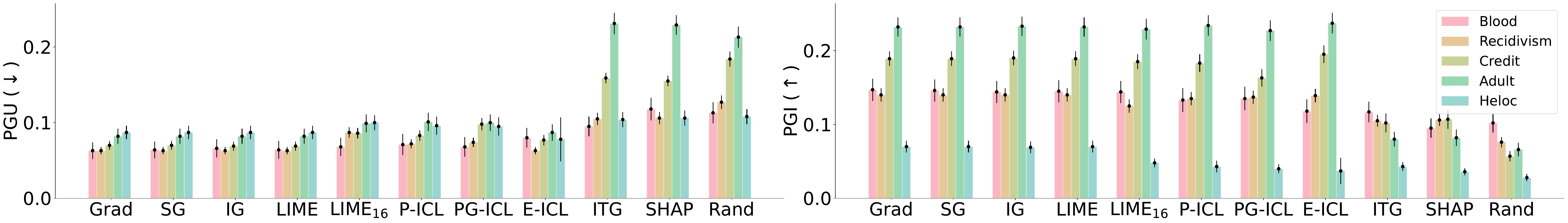}
    
    \caption{\small\looseness=-1 \textbf{Top:} FA (left) and RA (right) scores of explanations generated using post hoc explainers and GPT-4 (\pc, \pcg, and \EICL strategies) for an LR model. \textbf{Center:} PGU (left) and PGI (right) scores for a three-layer ANN model. \textbf{Bottom:} PGU (left) and PGI (right) scores for a six-layer ANN. On average, across four datasets and three predictive models, ICL strategies demonstrate non-trivial post hoc explanation capabilities: \EICL explanations (with in-context examples selected from LIME) match the faithfulness of gradient-based/LIME methods; 
    \pc and \pcg explanations achieve more faithful scores than ITG and SHAP methods, while demonstrating similar faithfulness to LIME$_{16}$ (\ie given the same number of input perturbations).}
    \label{fig:main-lr}
\end{figure*}



\begin{figure*}[h]
    \centering
    \includegraphics[width=0.99\linewidth]{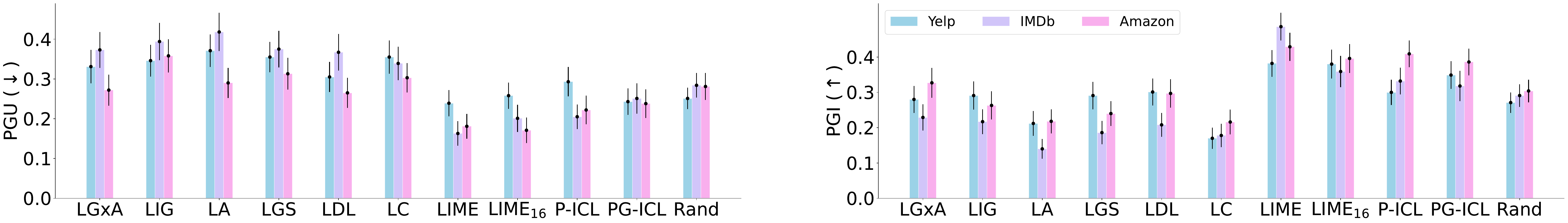}
    \vspace{-0.1in}
    \caption{\small
        \looseness=-1 PGU-text (left) and PGI-text (right) faithfulness scores of the sentiment classifier's explanations generated using post hoc explainers and LLMs (\pc and \pcg strategies) for Yelp, IMDb, and Amazon review datasets. Across three datasets, ICL strategies consistently demonstrate non-trivial post hoc explanation capabilities, achieving higher faithfulness than gradient-based layer-wise attributions, and approaching the performance of current state-of-the-art (LIME).
    }
    \label{fig:main-ann-xl}
    \vspace{-0.1in}
\end{figure*}

\begin{figure*}[h]
    \begin{flushleft}
        \footnotesize
        \hspace{2.9cm}\textsc{Logistic Regression}\hspace{5.0cm}\textsc{Neural Network}
    \end{flushleft}
    \vspace{-0.05in}
    \centering
    \includegraphics[width=0.99\linewidth]{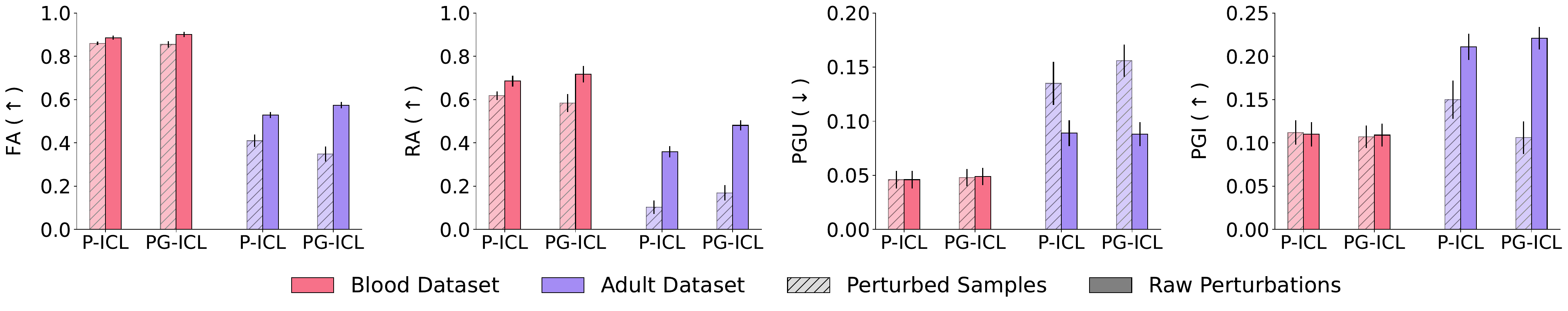}
    \vspace{-0.1in}
    \caption{\small\textbf{ICL set representation:} Faithfulness explanation performance of \pc and \pcg using the perturbed samples ($\mathbf{x'}$) and the raw perturbations ($\delta$) as shown in Fig.~\ref{fig:input_format} in the prompt for LR (left) and ANN (right) models. On average, across both prompting strategies and \textcolor{red!90!black}{Blood} and \textcolor{blue!90!black}{Adult} datasets, we find that generating ICL samples using the raw perturbation format results in significantly better faithfulness performance across all four metrics.}
    \label{fig:perturbation}
\end{figure*}

\begin{figure*}[h]
    \begin{flushleft}
        \footnotesize
        \hspace{2.9cm}\textsc{Logistic Regression}\hspace{5.0cm}\textsc{Neural Network}
    \end{flushleft}
    \vspace{-0.05in}
    \centering
    \includegraphics[width=0.99\linewidth]{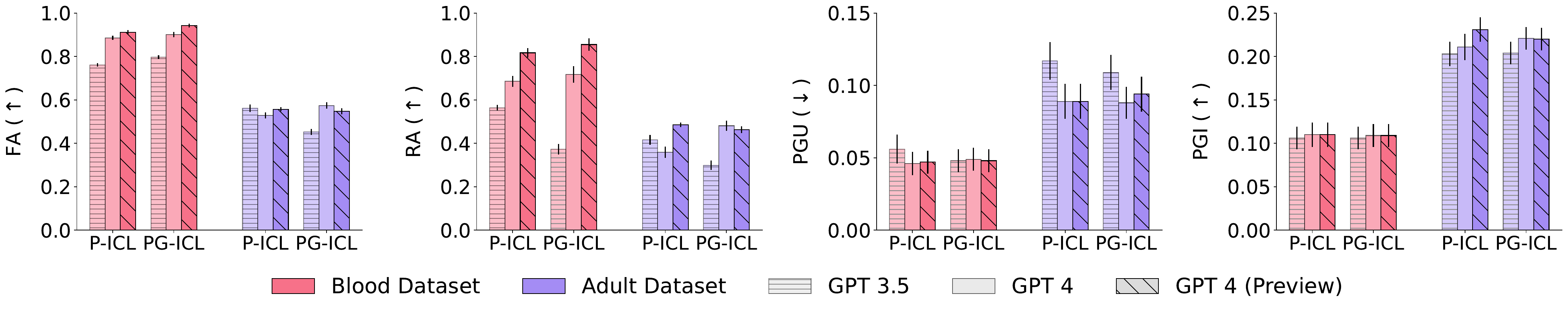}
    \vspace{-0.1in}
    \caption{\small\looseness=-1 \textbf{Choice of LLMs:} Faithfulness explanation performance of \pc and \pcg prompting strategies on \textcolor{red!90!black}{\blood} and \textcolor{blue!90!black}{\adult} datasets for different \textsc{Gpt} models. On average, across the LR (left) and ANN (right) models and both datasets, the \gptfournew explanations outperform the \gptthree and \gptfour models.}
    \label{fig:compare-pc-pcg-gpt-lr-ann}
\end{figure*}
\subsection{Results}
\label{sec:results}
Next, we discuss our results that answer key questions highlighted at the beginning of this section about LLMs as post hoc explainers (Q1-Q3).

\looseness=-1
\xhdr{1) ICL strategies can identify important tabular features} We compare the proposed prompting-based LLM explanation strategies, namely \pc, \pcg, and \EICL to existing post hoc explainers on the task of identifying important features for understanding different predictive models for tabular and text classification datasets. For the ANN model, the LLM-based explanations perform on par with post hoc explainers (despite having white-box access to the underlying predictive model or training surrogate linear model). 
We observe that LLM explanations, on average, achieve 50.13\% lower PGU and 144.82\% higher PGI than ITG, SHAP, and random baselines for larger datasets (more number of features) like Adult and Credit compared to 27.49\% lower PGU and 20.67\% higher PGI for Blood and Recidivism datasets. 
The improved performance of LLM explanations in datasets with more features suggests that LLMs may be better suited for handling the complexities of larger datasets.
While the LLM prompting strategies achieve competitive PGU and PGI scores across different datasets for ANN models, the \pcg strategy, on average across four datasets, achieves higher FA and RA scores than \pc for the LR model (Fig.~\ref{fig:main-lr}). 
Moreover, on average, \pcg achieves 6.89\% higher FA and 16.43\% higher RA across datasets compared to \pc. We find that gradient-based methods and LIME achieve almost perfect FA and RA scores as they can get accurate model gradients and approximate the model behavior with high precision. Interestingly, the LLM-based explanations perform better than ITG, SHAP, and Random baseline methods, even for a linear model.

\looseness=-1 Additionally, for the \EICL prompting strategy, LLM-augmented explainers achieve similar faithfulness to their vanilla counterparts. 
The results show that LLMs generate explanations that achieve faithfulness performance on par with those generated using post hoc explanation methods for LR and ANN predictive models across all five datasets (Fig.~\ref{fig:compas-lr}; see Table \ref{tab:exp_based_icl} for complete results) and four evaluation metrics. 
We demonstrate that very few in-context examples ($n_{\text{ICL}}=4$) are sufficient to make the LLM mimic the behavior of any post hoc explainer and generate faithful explanations, suggesting the effectiveness of LLMs as an explanation method, indicating that LLMs can effectively utilize their inherent capabilities to maintain the faithfulness of explanations in line with traditional methods.

\xhdr{2) ICL strategies identify important words for sentiment classifiers} Natural language is one domain of particular interest when assessing the capabilities of LLMs as explainers of other predictive models. When providing examples of removed words and corresponding changes in sentiment classification, we find that GPT-4 is able to consistently identify a top-3 most important wordset that achieves higher faithfulness than gradient-based layer-wise attributions (LGxA, LIG, LA, LGS, LDL, and LC), according to the modified PGU/PGI metrics described in Sec.~\ref{subsec:data_exps}. While LIME yields more faithful top-3 wordsets on average, we do observe cases where GPT-4 outperforms LIME-16, \eg a 35\% vs 30\% increase in PGI-text over the random baseline on the Amazon dataset, using the P-ICL prompting strategy. This approaches the gold standard of LIME at around 41\%, indicating promising capabilities given that a) only 16 examples are provided to the LLM, b) our strategies use solely the prompt in order to derive explanations (\ie ICL rather than finetuning). PG-ICL achieves higher faithfulness than P-ICL for the Yelp dataset (roughly a 30\% vs 10\% increase over random), whereas P-ICL, on average, yields better increases on the Amazon and IMDb datasets (around 14\% vs 9\%, and 35\% vs 27\%, respectively).

\looseness=-1
\looseness=-1

\xhdr{3) Ablation study} Here, we show how prompt modifications and choice of LLMs affect explanation faithfulness.

\xhdr{a) ICL set representation} \textit{Does the choice between the raw perturbation ($\delta$) and the perturbed sample ($\mathbf{x'}$) affect faithfulness?} On average, across the \blood and \adult datasets for both LR and ANN (Fig. ~\ref{fig:perturbation}), our results show that using raw perturbation in the prompts significantly aids LLMs in discerning the most important features. We find that providing only the raw perturbation bypasses the LLM's need to internally compute the difference \textit{w.r.t.} the test sample and this relational perspective allows the LLM to focus directly on variations in input and output.


\xhdr{b) Choice of LLMs} \textit{How do different LLMs impact faithfulness?} Here, we perform an ablation using different models from the \textsc{Gpt} family, \textit{viz.} \gptthree, \gptfour, and \gptfournew. Our results in Fig.~\ref{fig:compare-pc-pcg-gpt-lr-ann} show that faithfulness performance, on average, improves with LLM's capabilities. In particular, we observe that \gptfournew achieves better FA, RA, and PGI scores across both datasets and prompting strategies. We attribute this improvement to the LLM's updated knowledge and ability to reduce the ``\textit{laziness}'' cases, where the model doesn’t complete a task.

Please refer to Appendix~\ref{app:results} for additional ablation results on the impact of enforcing CoT, using context in the prompting template, different ICL set sizes ($n_{\text{ICL}}=\{4, 8, 12, 16, 32\}$), and other open-sourced LLMs like Llama and Mixtral. We detail this ablation in the Appendix due to space constraints. 
\section{Conclusion}
\label{sec:conclusion}
\looseness=-1 We introduce a novel framework, \emph{In-Context Explainers}, and explore the potential of using LLMs as post hoc explainers. To this end, we propose three prompting strategies --- \vanillaExp, \instructionICL, and \iclExp --- leveraging the context, dataset, and varying levels of instructions to generate explanations using LLMs for other predictive models. We conducted many experiments to evaluate LLM-generated explanations using eight datasets. Our results across different prompting strategies highlight that LLMs can generate faithful explanations, similar to post hoc explainers. Our work paves the way for several exciting future directions in explainable artificial intelligence (XAI) to explore LLM-based explanations.

\printbibliography
\appendix
\newpage
\section{Appendix: Additional results and Experimental details}
\label{app:appendix}

\subsection{Additional Experimental Details}
\label{app:details}
\xhdr{Bad replies} The total number of occurrences in which either the LLM didn't follow our requested response format or it replied with \textit{``I don’t have enough information to determine the most important features''} is detailed in Table ~\ref{tab:bad-replies}.

\begin{table*}[h]
\centering
\setlength{\tabcolsep}{0.9pt}
\renewcommand{\arraystretch}{0.9}
\caption{
Percentage of `\textit{bad replies}' produced by \gptfour across all datasets, models, and prompting strategies (\pc, \pcg, and \EICL under default prompt settings). Bad replies include refusal to answer or failure to return answers in a parseable format. Here, Rec. stands for Recidivism.
}
\footnotesize
\begin{tabular}{l@{\hspace{20pt}}l@{\hspace{20pt}}cccc@{\hspace{20pt}}cccc@{\hspace{20pt}}c}
\toprule
& \multicolumn{1}{l}{Method} & \multicolumn{4}{c}{LR} & \multicolumn{4}{c}{ANN} &  \\ & & \multicolumn{1}{l}{Blood} & \multicolumn{1}{l}{Rec.} & \multicolumn{1}{l}{Credit} & \multicolumn{1}{l}{Adult} & \multicolumn{1}{l}{Blood} & \multicolumn{1}{l}{Reci.} & \multicolumn{1}{l}{Credit} & \multicolumn{1}{l}{Adult} & \multicolumn{1}{l}{Mean percentage}      \\ \midrule
\multirow{3}{*}{GPT-3.5} & P-ICL & 0\%  & 1\%  & 2\%  & 1\%  & 0\%  & 0\%  & 0\%  & 1\% &0.62\%\\
& PG-ICL & 0\%  & 0\%  & 0\%  & 0\%  & 0\%  & 0\%  & 0\%  & 0\% &0.00\%\\
& E-ICL & 0\%  & 0\%  & 0\%  & 0\%  & 0\%  & 0\%  & 0\%  & 0\% &0.00\%\\
\midrule
\multirow{3}{*}{GPT-4} & P-ICL & 0\%  & 0\%  & 3\%  & 18\%  & 0\%  & 0\%  & 3\%  & 14\% &4.75\%\\
& PG-ICL & 0\%  & 0\%  & 0\%  & 0\%  & 0\%  & 0\%  & 0\%  & 0\% &0.00\%\\
& E-ICL & 0\%  & 0\%  & 0\%  & 0\%  & 0\%  & 0\%  & 0\%  & 0\% &0.00\%\\
\midrule
\multirow{3}{*}{GPT-4-Preview} & P-ICL & 0\%  & 0\%  & 3\%  & 2\%  & 0\%  & 0\%  & 3\%  & 2\% &1.25\%\\
& PG-ICL & 0\%  & 0\%  & 0\%  & 4\%  & 0\%  & 0\%  & 0\%  & 1\% &0.62\%\\
& E-ICL & 0\%  & 0\%  & 4\%  & 18\%  & 0\%  & 0\%  & 0\%  & 1\% &2.88\%\\
\bottomrule
\end{tabular}
\label{tab:bad-replies}
\end{table*}

\xhdr{Datasets}

The \textbf{Blood} dataset \cite{yeh2009knowledge} comprises of 4 attributes of 748 donors to a blood transfusion service from Taiwan. The task is to determine whether patients return for another donation.

The \textbf{Recidivism} dataset \cite{compas} has criminal records and demographics features for 6,172 defendants released on bail at U.S state courts during 1990-2009. The task is to classify defendants into bail (unlikely to commit a violent crime if released) vs. no bail (likely to commit one).

The \textbf{\credit} dataset \cite{default} includes financial and demographic data from credit card users at a bank, such as age, gender, education, marital status, credit limit, payment history, and bill amounts for several months. The primary task is to predict whether an individual will default on their payment.

The \textbf{Adult Income} dataset \cite{adult} contains demographic (e.g., age, race, and gender), education (degree), employment (occupation, hours-per week), personal (marital status, relationship), and financial (capital gain/loss) features for 45,222 individuals. The task is to predict whether an individual’s income exceeds \$50K per year vs. not.

The \textbf{HELOC} dataset \cite{heloc}  comprises of financial (e.g., total number of trades, average credit months in file) attributes from anonymized applications submitted by 9,871 real homeowners. A HELOC (Home Equity Line of Credit) is a line of credit typically offered by a bank as a percentage of home equity. The task is to  predict whether applicants will repay their HELOC within 2 years.

The \textbf{Amazon reviews} dataset contains textual reviews from users describing their experiences with products. Each review includes a rating provided by the user. The task involves predicting the sentiment (positive or negative) of each review based on its content. 

The \textbf{IMDb} dataset consists of movie reviews taken from the Internet Movie Database website. Each entry includes a textual review along with an associated sentiment label. The task involves predicting the sentiment (positive or negative) of each review based on its content. 

The \textbf{Yelp} dataset comprises user-generated reviews from Yelp, an online platform where individuals review various businesses and services. It contains reviews along with their corresponding star ratings, ranging from 1 to 5. The task involves predicting the sentiment (positive or negative) of each review based on its content.

\xhdr{Architecture details of ANNs} For the tabular dataset classsifiers, we used two different neural network models in our experiments: \textsc{ANN-L}, which has three hidden layers of size 64, 32, and 16, using \textsc{ReLU} for the hidden layers and \textsc{Softmax} for the output, and \textsc{ANN-XL},  which has six hidden layers of size 512, 256, 128, 64, 32, and 16 using \textsc{ReLU} for the hidden layers and \textsc{Softmax} for the output.

\xhdr{Architecture details of the Transformer} For the sentiment classification task, we use a transformer model with a token embedding layer to transform a tokenized sentence into real-dimensional vectors. The embeddings are added with sinusoidal positional encodings and then passed through a transformer encoder, configured with a single layer and four attention heads. The transformer also was trained with a dropout layer with $p=0.5$, followed by a linear binary-classification output layer and a softmax activation.

\xhdr{LLM perturbation hyperparameters} We use the LLM's top-$k$ features to calculate explanation faithfulness using four evaluation metrics. For calculating PGU and PGI metrics, we use perturbation mean $\mu_{PG}{=}0$, standard deviation $\sigma_{PG}{=}0.1$, and the number of perturbed samples $m_{PG}{=}10,000$. We follow the default hyperparameters from OpenXAI for generating explanations from standard post hoc explainers.

\xhdr{Metrics} We follow ~\cite{agarwal2022openxai} and used their evaluation metrics in our work. Below, we provide their respective definitions.

\emph{a) Feature Agreement (FA)} metric computes the fraction of top-$K$ features that are common between a given post hoc explanation and the corresponding ground truth explanation.

\emph{b) Rank Agreement (RA)} metric measures the fraction of top-$K$ features that are not only common between a given post hoc explanation and the corresponding ground truth explanation, but also have the same position in the respective rank orders.

\emph{c) Prediction Gap on Important feature perturbation (PGI)} metric measures the difference in prediction probability that results from perturbing the features deemed as influential by a given post hoc explanation.

\emph{d) Prediction Gap on Unimportant feature perturbation (PGU)} which measures the difference in prediction probability that results from perturbing the features deemed as unimportant by a given post hoc explanation. 

For a given instance $\mathbf{x}$, we first obtain the prediction probability $\hat{y}$ output by the underlying model $f$, \ie $\hat{y}=f(\mathbf{x})$. Let $e_{\mathbf{x}}$ be an explanation for the model prediction of $\mathbf{x}$. In the case of PGU, we then generate a perturbed instance $\mathbf{x}'$ in the local neighborhood of $\mathbf{x}$ by holding the top-$k$ features constant, and slightly perturbing the values of all the other features by adding a small amount of Gaussian noise. In the case of PGI, we generate a perturbed instance $\mathbf{x}'$ in the local neighborhood of $\mathbf{x}$ by slightly perturbing the values of the top-$k$ features by adding a small amount of Gaussian noise and holding all the other features constant. Finally, we compute the expected value of the prediction difference between the original and perturbed instances as:

\begin{equation}
    \textup{PGI}(\mathbf{x}, f, e_{\mathbf{x}}, k) = \mathbb{E}_{\mathbf{x}' \sim \textnormal{perturb($\mathbf{x}$, $e_{\mathbf{x}}$, top-$K$)}}[|\hat{y} - f(\mathbf{x}')|],
    \label{eq:pred_preserve}
\end{equation}

\begin{equation}
    \textup{PGU}(\mathbf{x}, f, e_{\mathbf{x}}, k) = \mathbb{E}_{\mathbf{x}' \sim \textnormal{perturb($\mathbf{x}$, $e_{\mathbf{x}}$, non top-$K$)}}[|\hat{y} - f(\mathbf{x}')|],
    \label{eq:pred_preserve_inv}
\end{equation}

where perturb($\cdot$) returns the noisy versions of $\mathbf{x}$ as described above.

For text datasets, we adapt the prediction gap metrics to assess explanation faithfulness more appropriately for natural language inputs:

\emph{e) Prediction Gap on Important word perturbation (PGI-text)} and \emph{f) Prediction Gap on Unimportant word perturbation (PGU-text)} measure the impact of the presence or absence of specific words identified as mportant or unimportant in the text classifier's decision. Unlike tabular datasets where perturbations involve adding noise, in text datasets, perturbations involves selectively removing words based on the post hoc explainer's importance ranking.

For a given text instance $\mathbf{x}$ represented by a sentence, we first obtain the classifier's prediction probability $\hat{y} = f(\mathbf{x})$. Let $e_{\mathbf{x}}$ denote an explanation that identifies the importance of words within $\mathbf{x}$. In the case of PGU-text, we generate a perturbed version $\mathbf{x}'$ by removing words not highlighted as top-$K$ important, simulating the omission of supposedly unimportant words. Conversely, for PGI-text, $\mathbf{x}'$ is formed by removing the top-$K$ words deemed important, to observe how their absence affects the prediction:

\begin{equation}
\textup{PGI-text}(\mathbf{x}, f, e_{\mathbf{x}}, K) = |\hat{y} - f(\mathbf{x}')|
\label{eq}
\end{equation}

\begin{equation}
\textup{PGU-text}(\mathbf{x}, f, e_{\mathbf{x}}, K) = |\hat{y} - f(\mathbf{x}')|
\label{eq}
\end{equation}

The Area Under the Curve (AUC) of these metrics across varying $k$ captures the overall faithfulness of the model's explanations.

\xhdr{Hyperparameters for XAI methods} Below, we provide the values for all hyperparameters of the explanation methods used in our experiments.

\xhdr{a) LIME}
kernel\_width           = 0.75;~
std\_LIME               = 0.1;~
mode                   = `\textit{tabular}';~
sample\_around\_instance = True;~
n\_samples\_LIME         = 1000 or 16;~
discretize\_continuous  = False

\xhdr{b) Grad}
absolute\_value = True

\xhdr{c) Smooth grad}
n\_samples\_SG = 100;~
std\_SG       = 0.005

\xhdr{d) Integrated gradients}
method             = `\textit{gausslegendre}';~
multiply\_by\_inputs = False;~
n\_steps            = 50

\xhdr{e) SHAP}
n\_samples = 500

\xhdr{f) \LayerIntGrad}
method = `\textit{gausslegendre}';~
n\_steps = 500

\xhdr{g) \LayerGradShap}
n\_samples = 5;~
stdevs = 0.0

\xhdr{h) \LayerCond}
method = `\textit{gausslegendre}';~
n\_steps = 50

\subsection{Additional Results}
\label{app:results}
Here, we include additional and detailed results of the experiments discussed in Sec.~\ref{sec:experiments}.

\begin{figure*}[h]
    \begin{flushleft}
        \footnotesize
        \hspace{2.2cm}\textsc{Logistic Regression}\hspace{5.8cm}\textsc{Neural Network}
    \end{flushleft}
    \centering
    \includegraphics[width=0.99\linewidth]{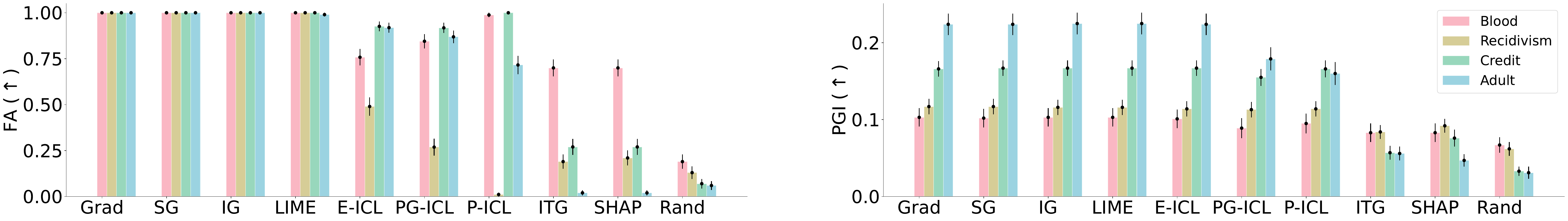}
    \vspace{-0.1in}
    \caption{\looseness=-1
    FA scores for LR (left) and PGI scores for ANN (right) for identifying the top-$k=1$ most important feature using post hoc explainers and LLMs (\pc, \pcg, and \EICL strategies).
    For both models, the LLM explanations generally perform on par with gradient-based methods across most datasets, with notable outlier dips in performance for the \compas dataset, indicating sensitivity to the dataset's characteristics.}
    \label{fig:top1-lr-ann}
\end{figure*}

\xhdr{Identifying top-$\mathbf{k{=}1}$ feature}~To demonstrate the LLM's capability in identifying the most important feature, we show the faithfulness performance of generated explanations across four datasets. 
In particular, for the LR model (in Table~\ref{tab:top1_lr} and Fig.~\ref{fig:top1-lr-ann}), we find significant variations in the feature agreement scores for the most important feature (top-$k=1$). 
Gradient-based methods like Grad, SG, and IG consistently achieved perfect FA scores across all datasets, demonstrating their reliability in identifying key features. 
In contrast, ITG and SHAP showed considerable variability in FA scores, particularly in the Recidivism and Adult datasets, with ITG recording as low as {0.190\std0.039} and {0.020\std0.014}, respectively. 
Among the LLM methods, \EICL performed notably well, achieving {0.490\std0.050} in Recidivism and {0.926\std0.027} in Credit, outperforming \pc and \pcg.  
However, \pc showed a significant disparity, with a low FA score of {0.011\std0.011} in Recidivism but high scores of {1.000\std0.000} in Credit and {0.988\std0.012} in Blood dataset. 
This lower performance in Recidivism is attributed to the LLM's approach to similarly important features, defaulting to alphabetical order in cases of near-equal importance.
\pcg generally showed more consistent and higher FA scores than \pc, particularly in Recidivism and Adult, \textbf{indicating its effectiveness in certain contexts by adding detailed guidance to assist in task execution}.

Our experiments for the ANN model (in Table~\ref{tab:top1_ann} and Fig.~\ref{fig:top1-lr-ann}) show that the top-$k{=}1$ features identified by our proposed prompting strategies achieve similar PGU and PGI scores to gradient-based and LIME post hoc explainers, showing the utility of LLMs when explaining non-linear ML models. In particular, we observe that for Recidivism, Credit, and Blood datasets, the \textbf{PGI scores obtained by \pc, \pcg, and \EICL are on par with that of all gradient-based and LIME explainers}. Overall, our results across four datasets and two ML models highlight the reliability of gradient-based methods and highlight the varying effectiveness of our prompting strategies.

\begin{figure*}
    \centering
    \includegraphics[width=0.99\linewidth]{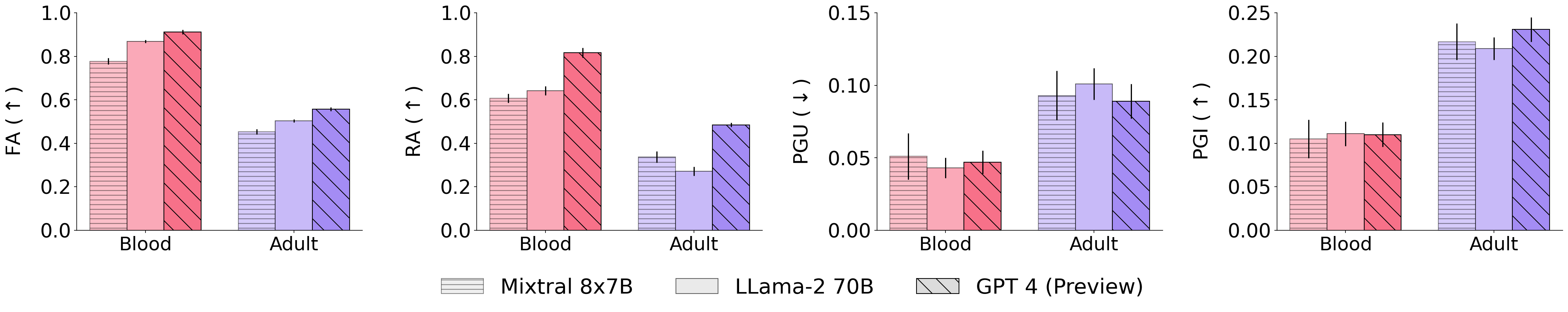}
    \vspace{-0.1in}
    \caption{Faithfulness explanation performance of \pc and \pcg prompting strategies on \textcolor{red!90!black}{\blood} and \textcolor{blue!90!black}{\adult} datasets for Llama-2 70B and Mixtral 8x7B models. Both achieve comparable performance on the evaluation metrics. However, GPT-4, on average across both datasets and metrics, outperforms the two open-sourced LLMs.}
    \label{fig:ablation-mixtral-llama}
\end{figure*}

\xhdr{Ablation: Open-Sourced LLMs} We conduct additional experiments using two open-source LLMs, namely Llama-2 70B and Mixtral 8x7B. Our results show that both models achieve comparable performance on the evaluation metrics. As expected, however, they do not exhibit capabilities as strong as GPT-4 (Fig.~\ref{fig:ablation-mixtral-llama}). This mirrors the trends shown in Fig.~\ref{fig:compare-pc-pcg-gpt-lr-ann}, demonstrating increasing post hoc explanation capabilities with increasing model size.

\begin{figure*}
    \begin{flushleft}
        \footnotesize
        \hspace{2.9cm}\textsc{Logistic Regression}\hspace{5.8cm}\textsc{Neural Network}
  \end{flushleft}
    \centering
    \includegraphics[width=0.99\linewidth]{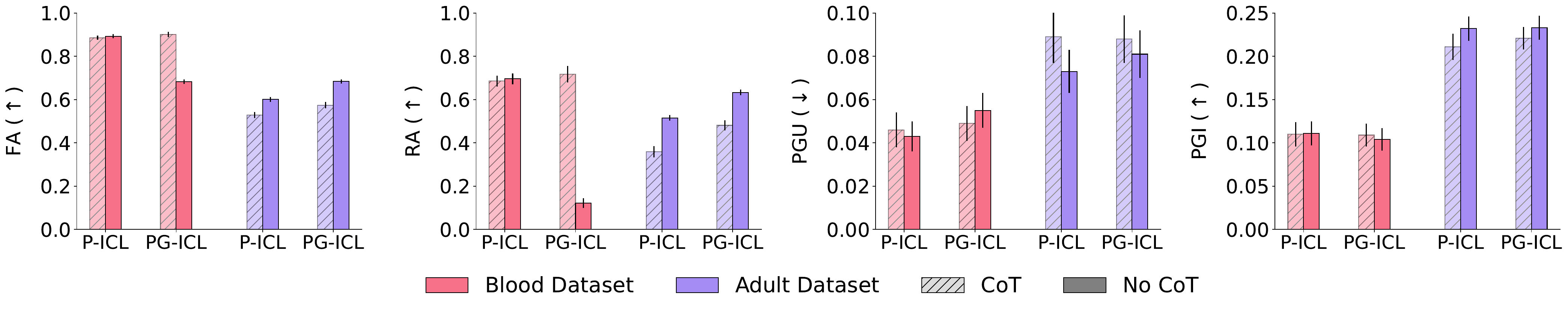}
    \vspace{-0.1in}
    \caption{Faithfulness explanation performance of \pc and \pcg prompting strategies on \textcolor{red!90!black}{\blood} and \textcolor{blue!90!black}{\adult} datasets w/ and w/o \textit{CoT} (\ie ``\textit{Think about the question}'') in the prompt template. 
    On average, across the LR (left) and ANN (right) models, both datasets, and prompting strategies, the effectiveness of \textit{CoT} varies but consistently enhances explanation faithfulness for the \adult dataset.}
    \label{fig:ablation-CoT-lr-ann}
\end{figure*}

\xhdr{Ablation: Impact of Chain-of-Thought (CoT)} \textit{Does incorporating a CoT approach before answering affect faithfulness?} We leverage the CoT capability of LLMs by including the phrase ``\textit{Think about the question}'' in the instructions. Our results show that for the \blood dataset, CoT's influence varies and does not consistently improve performance across models and prompting strategies (\pc and \pcg). Conversely, for the Adult dataset, omitting CoT generally enhances performance across both models and prompting strategies (Fig.~\ref{fig:ablation-CoT-lr-ann}), highlighting the significance of providing additional guidelines to the prompt in certain contexts.

\begin{figure*}
    \begin{flushleft}
        \footnotesize
        \hspace{2.9cm}\textsc{Logistic Regression}\hspace{5.8cm}\textsc{Neural Network}
  \end{flushleft}
    \centering
    \includegraphics[width=0.99\linewidth]{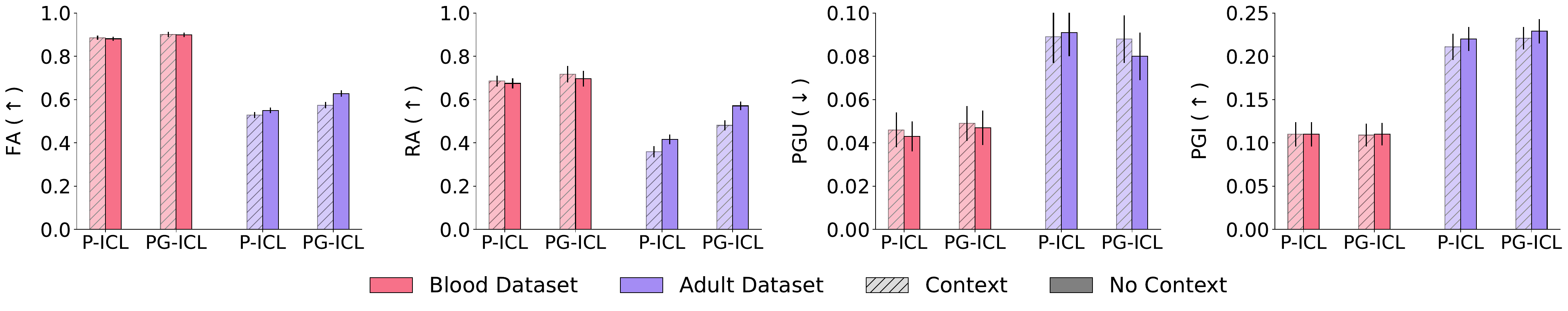}
    \vspace{-0.1in}
    \caption{Faithfulness explanation performance of \pc and \pcg prompting strategies on \textcolor{red!90!black}{\blood} and \textcolor{blue!90!black}{\adult} datasets w/ and w/o \textit{Context} in the prompt template. On average, across LR (left) and ANN (right) models and both datasets, we find that \textit{Context} doesn't have a significant impact on the faithfulness scores.
    }
    \label{fig:ablation-context-lr-ann}
\end{figure*}

\looseness=-1\xhdr{Ablation: Impact of setting the Context} \textit{How does the inclusion or exclusion of the prompt Context impact faithfulness?} Here, we take the original prompts in Figs.~\ref{fig:method1}-\ref{fig:method2} and remove the \textit{Context} from the template. On average, across two datasets and models, adding context doesn't have a significant impact on the faithfulness scores (Fig.~\ref{fig:ablation-context-lr-ann}), suggesting that the LLM can generate explanations using only the ICL set, question, and instructions.
This could imply that the LLMs are either inherently capable of generating faithful explanations without needing extra contextual clues, or that the specific type of context provided in these experiments does not contribute meaningfully to the explanation process.

\begin{figure*}[h]
    \centering
    \begin{flushleft}
        \footnotesize
        \hspace{3.5cm}\textsc{Blood Dataset}\hspace{6.2cm}\textsc{Adult Dataset}
    \end{flushleft}

    \includegraphics[width=0.99\linewidth]{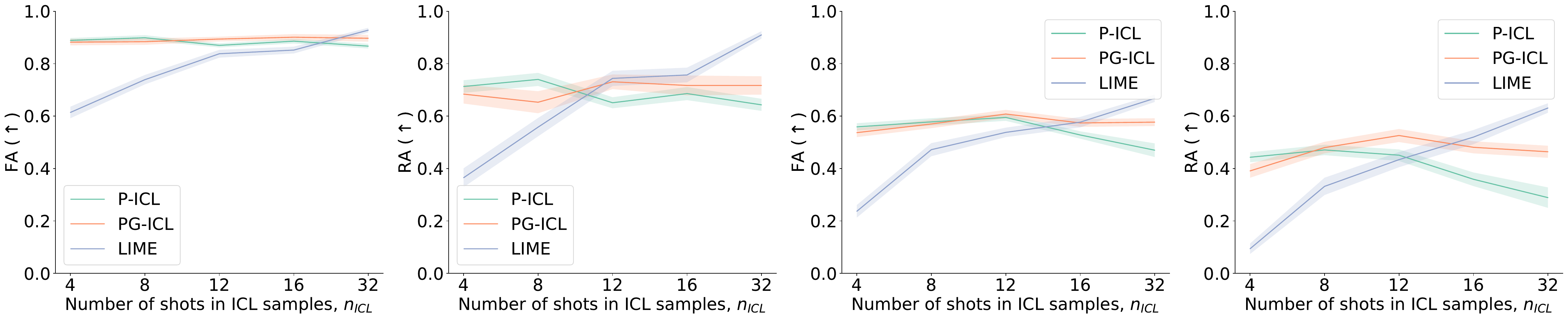}
    \caption{FA and RA performance of \pc and \pcg prompting strategies and LIME as we increase the number of ICL samples (analogous to neighborhood samples in LIME) for the LR model. In contrast to LIME, the faithfulness of LLM explanations across different metrics decreases for a higher number of ICL samples. This is likely due to limited capabilities of the LLM in tackling longer prompts, and/or its reluctance to analyze them (the number of successfully parsed replies decreases as we increase the number of samples).}
    \label{fig:compare-nshot}
\end{figure*}

\looseness=-1\xhdr{Ablation:  ICL set size} \textit{What impact do different ICL set sizes ($n_{\text{ICL}}$ = 4, 8, 12, 16, and 64) have on faithfulness?} Our ablation on the number of ICL samples (Fig.~\ref{fig:compare-nshot}) shows that fewer and larger numbers of ICL samples are not beneficial for LLMs to generate post hoc explanations. While fewer ICL samples provide insufficient information to the LLM to approximate the predictive behavior of the underlying ML model, a large number of ICL samples increases the input context, where the LLM struggles to retrieve relevant information from longer prompts, resulting in a decrease in the faithfulness of the explanations generated by LLMs. In contrast to LIME, the faithfulness of LLM explanations deteriorates upon increasing the number of ICL samples (analogous to the neighborhood of a given test sample).

\begin{figure*}[h]
    \centering
    \begin{flushleft}
        \footnotesize
        \hspace{2.5cm}\textsc{Logistic Regression}\hspace{4.2cm}\textsc{Neural Network}
  \end{flushleft}
    \begin{subfigure}{.24\textwidth}
      \centering
      \includegraphics[width=0.99\linewidth]{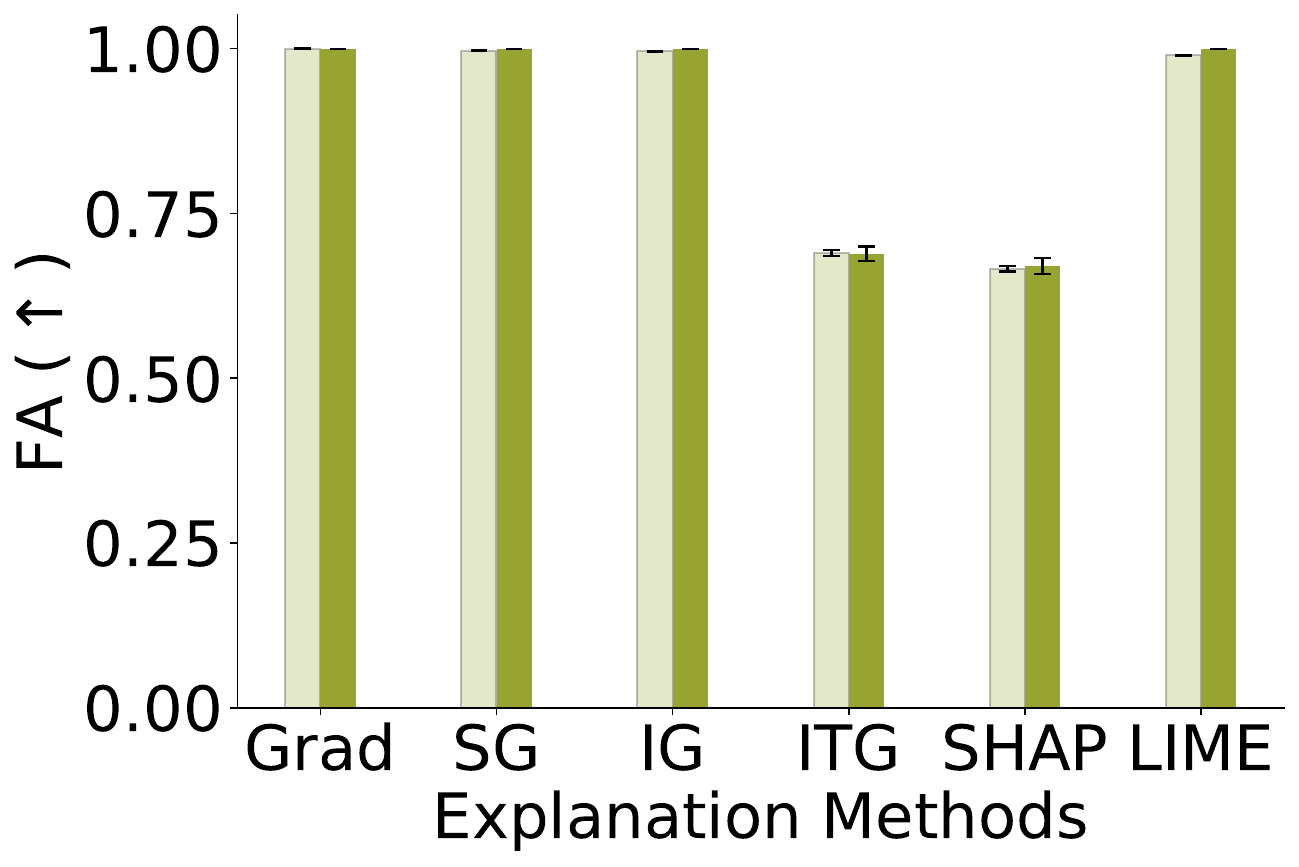}
      \label{fig:fa-compas-lr}
    \end{subfigure}
    \hfill
    \begin{subfigure}{.24\textwidth}
      \centering
      \includegraphics[width=0.99\linewidth]{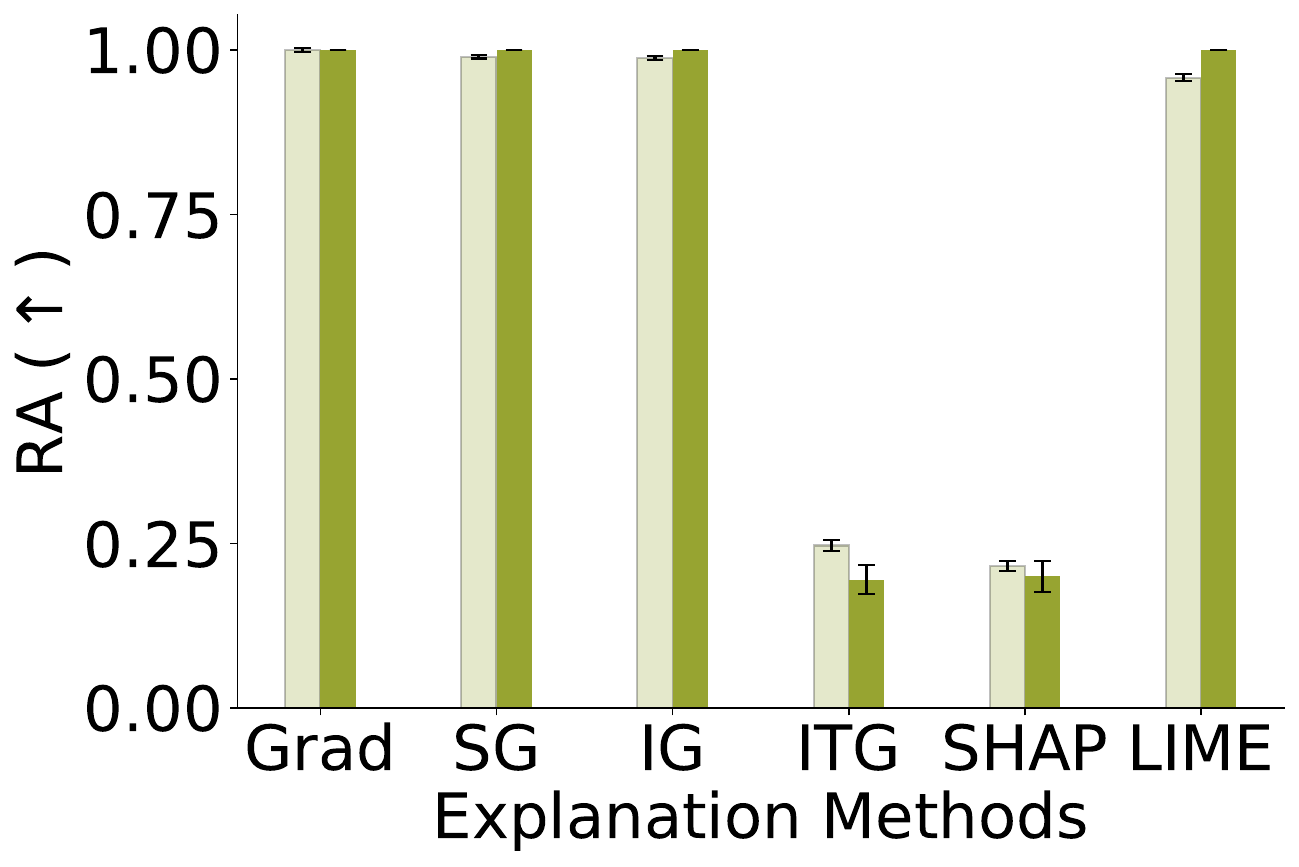}
      \label{fig:ra-compas-lr}
    \end{subfigure}
    \hfill
    \begin{subfigure}{.24\textwidth}
      \centering
      \includegraphics[width=0.99\linewidth]{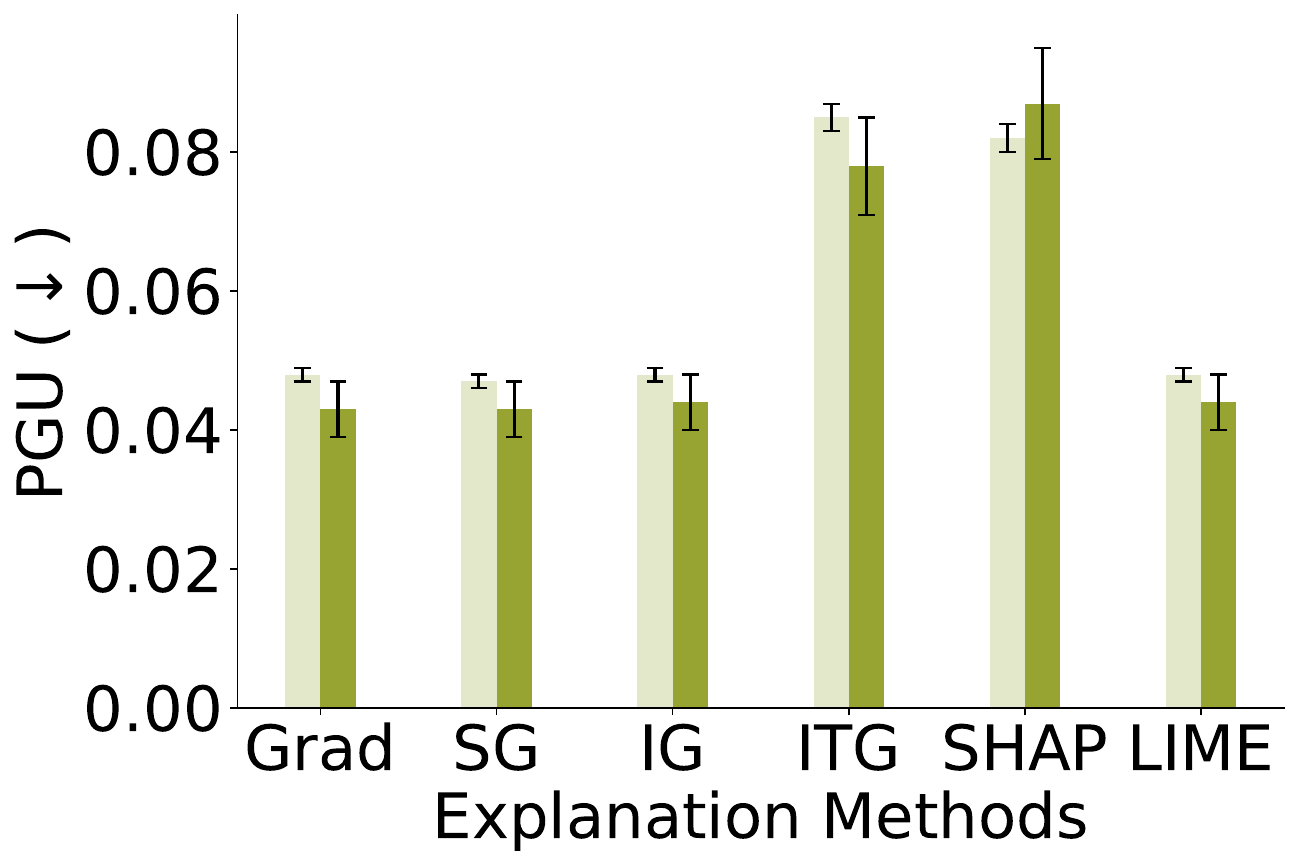}
      \label{fig:pgu-compas-ann-l}
    \end{subfigure}
    \hfill
    \begin{subfigure}{.24\textwidth}
      \centering
      \includegraphics[width=0.99\linewidth]{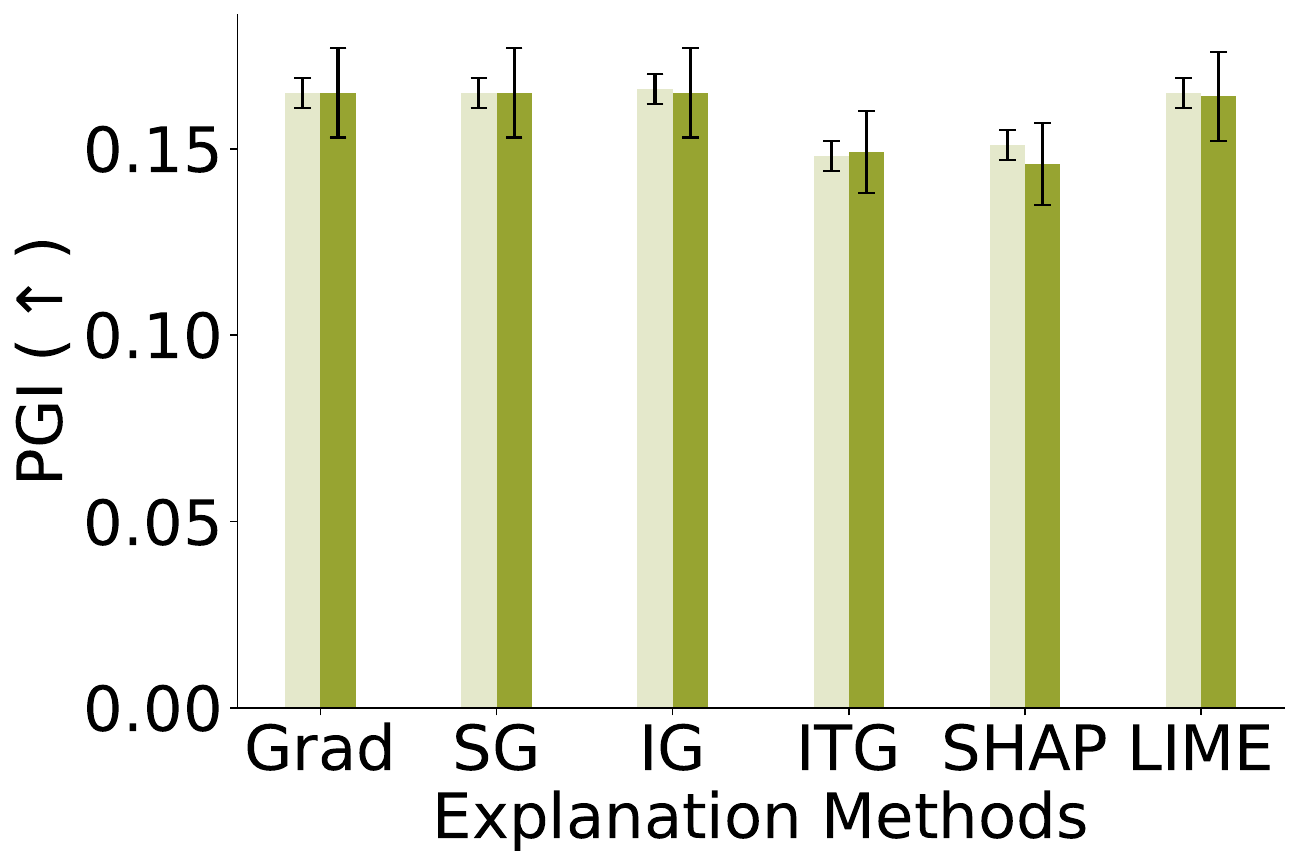}
      \label{fig:pgi-compas-ann-l}
    \end{subfigure}
    \begin{subfigure}{.54\textwidth}
      \centering
      \includegraphics[width=0.99\linewidth]{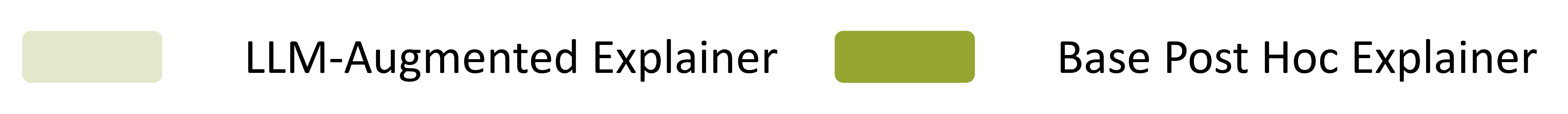}
      \label{fig:pgi-compas-lr}
    \end{subfigure}%
    \vspace{-0.25in}
    \caption{\looseness=-1 Faithfulness metrics for \EICL on the Recidivism dataset for six post hoc explainers and their LLM-augmented counterparts for a given LR (left) and ANN (right) model. LLM-augmented explanations achieve on-par performance \textit{w.r.t.} post hoc methods across all four metrics (see Table~\ref{tab:exp_based_icl} for complete results on all other datasets).
    Faithfulness metrics were computed for the top-$k$, $k$ being the number of features in each respective dataset.}
    \label{fig:compas-lr}
\end{figure*}

\begin{table}[h]
    \centering
    \vspace{-0.1in}
    \caption{
        Accuracy results for LR and ANN models across four datasets. 
        Best performance is shown in bold.
    }
    \vspace{-0.1in}
    {\begin{tabular}{lcc}
    \toprule
    {Dataset} & {LR} & {ANN} \\
    \toprule
    \begin{tabular}[l]{@{}l@{}}{\blood}\\{\compas}\\{\credit}\\{\adult}
    \end{tabular} &
    \begin{tabular}[c]{@{}c@{}}
    \textbf{70.59\%}\\
    {76.90\%}\\
    {87.37\%}\\
    77.37\%\\
    \end{tabular} &
    \begin{tabular}[c]{@{}c@{}}
    64.71\%\\
    76.90\%\\
    \textbf{88.34\%}\\
    \textbf{80.11\%}\\
    \end{tabular} \\
    \bottomrule
    \end{tabular}}
    \label{tab:performance}
\end{table}

\begin{table}[h]
    \centering
    \caption{
        Accuracy results for the sentiment-classification Transformer model across three datasets. 
    }
    {\begin{tabular}{lc}
    \toprule
    {Dataset} & {Transformer} \\
    \toprule
    \begin{tabular}[l]{@{}l@{}}{Amazon reviews}\\{IMDb}\\{Yelp}\\
    \end{tabular} &
    \begin{tabular}[c]{@{}c@{}}
    {72.50\%}\\
    {76.25\%}\\
    {71.25\%}\\
    \end{tabular} \\
    \bottomrule
    \end{tabular}}
    \label{tab:performance_transformer}
\end{table}

\begin{table*}[h]
\setlength{\tabcolsep}{1pt}
\renewcommand{\arraystretch}{1.1}
\footnotesize
\caption{Average faithfulness metric values and their standard errors for explanations across 100 test instances. 
It compares explanations from \vanillaExp and \instructionICL using \gptfour, six post hoc explanation methods, and a random baseline for LR and ANN predictions on four datasets.
For the LLM methods, we queried the LLM for the top-$k=5$ ($k=4$ for Blood) most important features and calculated each metric's area under the curve (AUC) for $k=3$ (where the AUC is calculated from $k=1$ to $k=3$). 
Arrows ($\uparrow$, $\downarrow$) indicate the direction of better performance. 
\label{tab:complete-icl-perturb-results}
}

\centering
\begin{tabular}{llcccc|cc}

	\toprule
	&                                                        & \multicolumn{4}{c}{LR}                                                                                                   & \multicolumn{2}{c}{ANN}                                                                           \\ \cline{3-8}
	Dataset                              & Method                                  & \multicolumn{1}{c}{FA ($\uparrow$)}      & \multicolumn{1}{c}{RA ($\uparrow$)}      & \multicolumn{1}{c}{PGU ($\downarrow$)}   & PGI ($\uparrow$)                         & \multicolumn{1}{c}{PGU ($\downarrow$)}   & PGI ($\uparrow$)                       \\ \toprule
	\multicolumn{1}{c}{}                 & Grad                                    & {1.000\std{0.000}}                       & {1.000\std{0.000}}         a             & {0.010\std{0.000}}                       & 0.042\std{0.000}                         & {0.060\std0.009}                         & 0.115\std0.013                         \\ 
	\multicolumn{1}{c}{}                 & SG                                      & {1.000\std0.000}                         & {1.000\std0.000}                         & {0.010\std0.000}                         & 0.042\std0.000                           & {0.060\std0.009}                         & 0.115\std0.013                         \\ 
	\multicolumn{1}{c}{}                 & IG                                      & {1.000\std0.000}                         & {1.000\std0.000}                         & {0.010\std0.000}                         & 0.042\std0.000                           & {0.061\std0.009}                         & 0.116\std0.013                         \\
	\multicolumn{1}{c}{}                 & ITG                                     & {0.722\std0.019}                         & {0.563\std0.037}                         & {0.019\std0.001}                         & 0.037\std0.001                           & {0.081\std0.010}                         & 0.100\std0.012                         \\ 
	\multicolumn{1}{c}{}                 & SHAP                                    & {0.723\std0.020}                         & {0.556\std0.037}                         & {0.019\std0.001}                         & 0.036\std0.001                           & {0.085\std0.011}                         & 0.098\std0.012                         \\ 
	\multicolumn{1}{c}{}                 & LIME                                    & {1.000\std0.000}                         & {1.000\std0.000}                         & {0.010\std0.000}                         & 0.042\std0.000                           & {0.061\std0.009}                         & 0.116\std0.013                         \\ 
	\multicolumn{1}{c}{}                 & Random                                  & {0.502\std0.022}                         & {0.232\std0.032}                         & {0.029\std0.001}                         & 0.026\std0.001                           & {0.091\std0.011}                         & 0.090\std0.012                         \\  
	\multicolumn{1}{c}{}                 & \cellcolor[HTML]{C0C0C0}\classicICL     & \cellcolor[HTML]{C0C0C0}{0.834\std0.013} & \cellcolor[HTML]{C0C0C0}{0.724\std0.023} & \cellcolor[HTML]{C0C0C0}{0.013\std0.000} & \cellcolor[HTML]{C0C0C0}{0.041\std0.000} & \cellcolor[HTML]{C0C0C0}{0.061\std0.010} & \cellcolor[HTML]{C0C0C0}{0.105\std0.013} \\ 
	{\multirow{-10}{*}{Blood}}           & \cellcolor[HTML]{C0C0C0}\instructionICL & \cellcolor[HTML]{C0C0C0}{0.875\std0.016} & \cellcolor[HTML]{C0C0C0}{0.742\std0.039} & \cellcolor[HTML]{C0C0C0}{0.013\std0.000} & \cellcolor[HTML]{C0C0C0}{0.041\std0.001} & \cellcolor[HTML]{C0C0C0}{0.064\std0.010} & \cellcolor[HTML]{C0C0C0}{0.104\std0.013} \\ \midrule
	                                     & Grad                                    & {1.000\std0.000}                         & {1.000\std0.000}                         & {0.059\std0.003}                         & 0.106\std0.005                           & {0.095\std0.008}                         & 0.149\std0.011                         \\  
	                                     & SG                                      & {1.000\std0.000}                         & {1.000\std0.000}                         & {0.059\std0.003}                         & 0.106\std0.005                           & {0.095\std0.008}                         & 0.149\std0.011                         \\  
	                                     & IG                                      & {1.000\std0.000}                         & {1.000\std0.000}                         & {0.059\std0.003}                         & 0.106\std0.005                           & {0.096\std0.008}                         & 0.149\std0.011                         \\  
	                                     & ITG                                     & {0.493\std0.021}                         & {0.214\std0.030}                         & {0.090\std0.005}                         & 0.078\std0.004                           & {0.129\std0.011}                         & 0.122\std0.010                         \\  
	                                     & SHAP                                    & {0.473\std0.023}                         & {0.217\std0.032}                         & {0.092\std0.005}                         & 0.076\std0.004                           & {0.130\std0.011}                         & 0.122\std0.010                         \\  
	                                     & LIME                                    & {1.000\std0.000}                         & {1.000\std0.000}                         & {0.059\std0.003}                         & 0.106\std0.005                           & {0.096\std0.008}                         & 0.149\std0.011                         \\  
	                                     & Random                                  & {0.308\std0.023}                         & {0.127\std0.024}                         & {0.101\std0.005}                         & 0.063\std0.005                           & {0.146\std0.011}                         & 0.092\std0.009                         \\ 
	                                     & \cellcolor[HTML]{C0C0C0}\classicICL     & \cellcolor[HTML]{C0C0C0}{0.746\std0.003} & \cellcolor[HTML]{C0C0C0}{0.088\std0.005} & \cellcolor[HTML]{C0C0C0}{0.061\std0.003} & \cellcolor[HTML]{C0C0C0}{0.105\std0.005} & \cellcolor[HTML]{C0C0C0}{0.096\std0.008} & \cellcolor[HTML]{C0C0C0}{0.147\std0.012} \\  
	\multirow{-10}{*}{Recidivism}        & \cellcolor[HTML]{C0C0C0}\instructionICL & \cellcolor[HTML]{C0C0C0}{0.756\std0.013} & \cellcolor[HTML]{C0C0C0}{0.275\std0.037} & \cellcolor[HTML]{C0C0C0}{0.065\std0.004} & \cellcolor[HTML]{C0C0C0}{0.103\std0.005} & \cellcolor[HTML]{C0C0C0}{0.099\std0.009} & \cellcolor[HTML]{C0C0C0}{0.146\std0.012} \\ \midrule
	                                     & Grad                                    & {1.000\std0.000}                         & {1.000\std0.000}                         & {0.065\std0.005}                         & 0.195\std0.009                           & {0.072\std0.008}                         & 0.173\std0.011                         \\  
	                                     & SG                                      & {1.000\std0.000}                         & {1.000\std0.000}                         & {0.065\std0.005}                         & 0.195\std0.009                           & {0.072\std0.008}                         & 0.172\std0.011                         \\  
	                                     & IG                                      & {1.000\std0.000}                         & {1.000\std0.000}                         & {0.065\std0.005}                         & 0.195\std0.009                           & {0.074\std0.008}                         & 0.172\std0.010                         \\  
	                                     & ITG                                     & {0.211\std0.026}                         & {0.157\std0.026}                         & {0.150\std0.006}                         & 0.106\std0.012                           & {0.155\std0.009}                         & 0.089\std0.011                         \\  
	                                     & SHAP                                    & {0.212\std0.026}                         & {0.161\std0.026}                         & {0.150\std0.006}                         & 0.107\std0.012                           & {0.150\std0.008}                         & 0.098\std0.012                         \\  
	                                     & LIME                                    & {0.988\std0.005}                         & {0.985\std0.007}                         & {0.065\std0.005}                         & 0.195\std0.009                           & {0.071\std0.008}                         & 0.173\std0.010                         \\  
	                                     & Random                                  & {0.173\std0.020}                         & {0.095\std0.020}                         & {0.185\std0.010}                         & 0.054\std0.006                           & {0.176\std0.011}                         & 0.053\std0.007                         \\ 
	                                     & \cellcolor[HTML]{C0C0C0}\classicICL     & \cellcolor[HTML]{C0C0C0}{0.622\std0.008} & \cellcolor[HTML]{C0C0C0}{0.604\std0.008} & \cellcolor[HTML]{C0C0C0}{0.077\std0.006} & \cellcolor[HTML]{C0C0C0}{0.192\std0.009} & \cellcolor[HTML]{C0C0C0}{0.080\std0.009} & \cellcolor[HTML]{C0C0C0}{0.171\std0.011} \\  
	\multirow{-10}{*}{\makecell{Credit}} & \cellcolor[HTML]{C0C0C0}\instructionICL & \cellcolor[HTML]{C0C0C0}{0.646\std0.014} & \cellcolor[HTML]{C0C0C0}{0.594\std0.020} & \cellcolor[HTML]{C0C0C0}{0.081\std0.007} & \cellcolor[HTML]{C0C0C0}{0.186\std0.009} & \cellcolor[HTML]{C0C0C0}{0.084\std0.009} & \cellcolor[HTML]{C0C0C0}{0.165\std0.011} \\ \midrule
	                                     & Grad                                    & {0.999\std0.001}                         & {0.999\std0.001}                         & {0.056\std0.006}                         & 0.221\std0.011                           & {0.081\std0.011}                         & 0.228\std0.014                         \\  
	                                     & SG                                      & {0.999\std0.001}                         & {0.999\std0.001}                         & {0.056\std0.006}                         & 0.221\std0.011                           & {0.080\std0.011}                         & 0.227\std0.014                         \\  
	                                     & IG                                      & {1.000\std0.000}                         & {1.000\std0.000}                         & {0.056\std0.006}                         & 0.221\std0.011                           & {0.082\std0.011}                         & 0.228\std0.014                         \\  
	                                     & ITG                                     & {0.385\std0.012}                         & {0.099\std0.019}                         & {0.215\std0.011}                         & 0.061\std0.007                           & {0.227\std0.014}                         & 0.075\std0.010                         \\  
	                                     & SHAP                                    & {0.387\std0.012}                         & {0.150\std0.020}                         & {0.215\std0.011}                         & 0.061\std0.007                           & {0.225\std0.014}                         & 0.075\std0.010                         \\  
	                                     & LIME                                    & {0.963\std0.012}                         & {0.953\std0.015}                         & {0.056\std0.006}                         & 0.221\std0.011                           & {0.078\std0.011}                         & 0.229\std0.014                         \\  
	                                     & Random                                  & {0.130\std0.017}                         & {0.053\std0.015}                         & {0.198\std0.012}                         & 0.054\std0.008                           & {0.213\std0.014}                         & 0.064\std0.010                         \\  
	                                     & \cellcolor[HTML]{C0C0C0}\classicICL     & \cellcolor[HTML]{C0C0C0}{0.541\std0.022} & \cellcolor[HTML]{C0C0C0}{0.450\std0.033} & \cellcolor[HTML]{C0C0C0}{0.086\std0.009} & \cellcolor[HTML]{C0C0C0}{0.197\std0.012} & \cellcolor[HTML]{C0C0C0}{0.110\std0.013} & \cellcolor[HTML]{C0C0C0}{0.197\std0.015} \\  
	\multirow{-10}{*}{Adult}             & \cellcolor[HTML]{C0C0C0}\instructionICL & \cellcolor[HTML]{C0C0C0}{0.669\std0.021} & \cellcolor[HTML]{C0C0C0}{0.622\std0.028} & \cellcolor[HTML]{C0C0C0}{0.075\std0.007} & \cellcolor[HTML]{C0C0C0}{0.210\std0.011} & \cellcolor[HTML]{C0C0C0}{0.107\std0.013} & \cellcolor[HTML]{C0C0C0}{0.208\std0.014} \\ \bottomrule
\end{tabular}
\end{table*}

\begin{table*}[h]
\centering
\setlength{\tabcolsep}{2.1pt}
\renewcommand{\arraystretch}{1.1}
\footnotesize
\caption{
Average faithfulness metric values and their standard errors for explanations across 100 test instances. 
It compares explanations from \iclExp using \gptfour and six post hoc methods for LR and ANN predictions on four datasets. Metrics were calculated for the top-$k$ features, with $k$ matching the dataset feature count. Arrows ($\uparrow$, $\downarrow$) denote the direction of improved performance.
}
\begin{tabular}{llcccc|cc}
\toprule
 &  & \multicolumn{4}{c}{LR} & \multicolumn{2}{c}{ANN} \\ \cline{3-8}
Dataset & Method & \multicolumn{1}{c}{FA ($\uparrow$)} & \multicolumn{1}{c}{RA ($\uparrow$)} & \multicolumn{1}{c}{PGU ($\downarrow$)} & PGI ($\uparrow$) & \multicolumn{1}{c}{PGU ($\downarrow$)} & PGI ($\uparrow$)\\
\toprule
\multicolumn{1}{c}{} & \cellcolor[HTML]{C0C0C0}LLM-Lime & \multicolumn{1}{c}{\cellcolor[HTML]{C0C0C0}1.000\std0.000} & \multicolumn{1}{c}{\cellcolor[HTML]{C0C0C0}0.978\std0.011} & \multicolumn{1}{c}{\cellcolor[HTML]{C0C0C0}0.000\std0.000} & \cellcolor[HTML]{C0C0C0}0.041\std0.001 & \multicolumn{1}{c}{\cellcolor[HTML]{C0C0C0}0.074\std0.009} & \cellcolor[HTML]{C0C0C0}0.099\std0.012 \\  
\multicolumn{1}{c}{} & Lime & \multicolumn{1}{c}{1.000\std0.000} & \multicolumn{1}{c}{1.000\std0.000} & \multicolumn{1}{c}{0.008\std0.000} & 0.043\std0.000 & \multicolumn{1}{c}{0.044\std0.006} & 0.121\std0.013 \\  
\multicolumn{1}{c}{} & \cellcolor[HTML]{C0C0C0}LLM-Grad & \multicolumn{1}{c}{\cellcolor[HTML]{C0C0C0}0.997\std0.003} & \multicolumn{1}{c}{\cellcolor[HTML]{C0C0C0}0.996\std0.004} & \multicolumn{1}{c}{\cellcolor[HTML]{C0C0C0}0.008\std0.000} & \cellcolor[HTML]{C0C0C0}0.043\std0.000 & \multicolumn{1}{c}{\cellcolor[HTML]{C0C0C0}0.058\std0.009} & \cellcolor[HTML]{C0C0C0}0.116\std0.012 \\  
\multicolumn{1}{c}{} & Grad & \multicolumn{1}{c}{1.000\std0.000} & \multicolumn{1}{c}{1.000\std0.000} & \multicolumn{1}{c}{0.008\std0.000} & 0.043\std0.000 & \multicolumn{1}{c}{0.044\std0.006} & 0.120\std0.013 \\  
\multicolumn{1}{c}{} & \cellcolor[HTML]{C0C0C0}LLM-SG & \multicolumn{1}{c}{\cellcolor[HTML]{C0C0C0}0.990\std0.006} & \multicolumn{1}{c}{\cellcolor[HTML]{C0C0C0}0.983\std0.011} & \multicolumn{1}{c}{\cellcolor[HTML]{C0C0C0}0.008\std0.000} & \cellcolor[HTML]{C0C0C0}0.043\std0.000 & \multicolumn{1}{c}{\cellcolor[HTML]{C0C0C0}0.055\std0.008} & \cellcolor[HTML]{C0C0C0}0.116\std0.012 \\  
\multicolumn{1}{c}{} & SG & \multicolumn{1}{c}{1.000\std0.000} & \multicolumn{1}{c}{1.000\std0.000} & \multicolumn{1}{c}{0.008\std0.000} & 0.043\std0.000 & \multicolumn{1}{c}{0.044\std0.006} & 0.120\std0.013 \\  
\multicolumn{1}{c}{} & \cellcolor[HTML]{C0C0C0}LLM-IG & \multicolumn{1}{c}{\cellcolor[HTML]{C0C0C0}0.989\std0.005} & \multicolumn{1}{c}{\cellcolor[HTML]{C0C0C0}0.982\std0.009} & \multicolumn{1}{c}{\cellcolor[HTML]{C0C0C0}0.008\std0.000} & \cellcolor[HTML]{C0C0C0}0.043\std0.000 & \multicolumn{1}{c}{\cellcolor[HTML]{C0C0C0}0.046\std0.007} & \cellcolor[HTML]{C0C0C0}0.120\std0.013 \\  
\multicolumn{1}{c}{} & IG & \multicolumn{1}{c}{1.000\std0.000} & \multicolumn{1}{c}{1.000\std0.000} & \multicolumn{1}{c}{0.008\std0.000} & 0.043\std0.000 & \multicolumn{1}{c}{0.044\std0.006} & 0.120\std0.013 \\  
\multicolumn{1}{c}{} & \cellcolor[HTML]{C0C0C0}LLM-Shap & \multicolumn{1}{c}{\cellcolor[HTML]{C0C0C0}0.684\std0.013} & \multicolumn{1}{c}{\cellcolor[HTML]{C0C0C0}0.401\std0.025} & \multicolumn{1}{c}{\cellcolor[HTML]{C0C0C0}0.020\std0.001} & \cellcolor[HTML]{C0C0C0}0.034\std0.001 & \multicolumn{1}{c}{\cellcolor[HTML]{C0C0C0}0.069\std0.009} & \cellcolor[HTML]{C0C0C0}0.102\std0.012 \\  
\multicolumn{1}{c}{} & Shap & \multicolumn{1}{c}{0.773\std0.014} & \multicolumn{1}{c}{0.516\std0.033} & \multicolumn{1}{c}{0.015\std0.001} & 0.038\std0.001 & \multicolumn{1}{c}{0.066\std0.009} & 0.107\std0.012 \\  
\multicolumn{1}{c}{} & \cellcolor[HTML]{C0C0C0}LLM-ITG & \multicolumn{1}{c}{\cellcolor[HTML]{C0C0C0}0.702\std0.013} & \multicolumn{1}{c}{\cellcolor[HTML]{C0C0C0}0.387\std0.029} & \multicolumn{1}{c}{\cellcolor[HTML]{C0C0C0}0.017\std0.001} & \cellcolor[HTML]{C0C0C0}0.036\std0.001 & \multicolumn{1}{c}{\cellcolor[HTML]{C0C0C0}0.069\std0.010} & \cellcolor[HTML]{C0C0C0}0.105\std0.012 \\  
\multicolumn{1}{l}{\multirow{-12}{*}{Blood}} & ITG & \multicolumn{1}{c}{0.774\std0.014} & \multicolumn{1}{c}{0.532\std0.034} & \multicolumn{1}{c}{0.014\std0.001} & 0.038\std0.001 & \multicolumn{1}{c}{0.063\std0.008} & 0.108\std0.012 \\ \midrule
 & \cellcolor[HTML]{C0C0C0}LLM-Lime & \multicolumn{1}{c}{\cellcolor[HTML]{C0C0C0}0.990\std0.001} & \multicolumn{1}{c}{\cellcolor[HTML]{C0C0C0}0.958\std0.005} & \multicolumn{1}{c}{\cellcolor[HTML]{C0C0C0}0.029\std0.001} & \cellcolor[HTML]{C0C0C0}0.115\std0.002 & \multicolumn{1}{c}{\cellcolor[HTML]{C0C0C0}0.048\std0.001} & \cellcolor[HTML]{C0C0C0}0.165\std0.004 \\  
 & Lime & \multicolumn{1}{c}{1.000\std0.000} & \multicolumn{1}{c}{1.000\std0.000} & \multicolumn{1}{c}{0.029\std0.002} & 0.116\std0.006 & \multicolumn{1}{c}{0.044\std0.004} & 0.164\std0.012 \\  
 & \cellcolor[HTML]{C0C0C0}LLM-Grad & \multicolumn{1}{c}{\cellcolor[HTML]{C0C0C0}0.997\std0.001} & \multicolumn{1}{c}{\cellcolor[HTML]{C0C0C0}0.990\std0.003} & \multicolumn{1}{c}{\cellcolor[HTML]{C0C0C0}0.029\std0.001} & \cellcolor[HTML]{C0C0C0}0.115\std0.002 & \multicolumn{1}{c}{\cellcolor[HTML]{C0C0C0}0.048\std0.001} & \cellcolor[HTML]{C0C0C0}0.165\std0.004 \\  
 & Grad & \multicolumn{1}{c}{1.000\std0.000} & \multicolumn{1}{c}{1.000\std0.000} & \multicolumn{1}{c}{0.029\std0.002} & 0.116\std0.006 & \multicolumn{1}{c}{0.043\std0.004} & 0.165\std0.012 \\  
 & \cellcolor[HTML]{C0C0C0}LLM-SG & \multicolumn{1}{c}{\cellcolor[HTML]{C0C0C0}0.997\std0.001} & \multicolumn{1}{c}{\cellcolor[HTML]{C0C0C0}0.990\std0.003} & \multicolumn{1}{c}{\cellcolor[HTML]{C0C0C0}0.029\std0.001} & \cellcolor[HTML]{C0C0C0}0.115\std0.002 & \multicolumn{1}{c}{\cellcolor[HTML]{C0C0C0}0.047\std0.001} & \cellcolor[HTML]{C0C0C0}0.165\std0.004 \\  
 & SG & \multicolumn{1}{c}{1.000\std0.000} & \multicolumn{1}{c}{1.000\std0.000} & \multicolumn{1}{c}{0.029\std0.002} & 0.116\std0.006 & \multicolumn{1}{c}{0.043\std0.004} & 0.165\std0.012 \\  
 & \cellcolor[HTML]{C0C0C0}LLM-IG & \multicolumn{1}{c}{\cellcolor[HTML]{C0C0C0}0.996\std0.001} & \multicolumn{1}{c}{\cellcolor[HTML]{C0C0C0}0.988\std0.003} & \multicolumn{1}{c}{\cellcolor[HTML]{C0C0C0}0.029\std0.001} & \cellcolor[HTML]{C0C0C0}0.115\std0.002 & \multicolumn{1}{c}{\cellcolor[HTML]{C0C0C0}0.048\std0.001} & \cellcolor[HTML]{C0C0C0}0.166\std0.004 \\  
 & IG & \multicolumn{1}{c}{1.000\std0.000} & \multicolumn{1}{c}{1.000\std0.000} & \multicolumn{1}{c}{0.029\std0.002} & 0.116\std0.006 & \multicolumn{1}{c}{0.044\std0.004} & 0.165\std0.012 \\  
 & \cellcolor[HTML]{C0C0C0}LLM-Shap & \multicolumn{1}{c}{\cellcolor[HTML]{C0C0C0}0.666\std0.004} & \multicolumn{1}{c}{\cellcolor[HTML]{C0C0C0}0.216\std0.008} & \multicolumn{1}{c}{\cellcolor[HTML]{C0C0C0}0.057\std0.001} & \cellcolor[HTML]{C0C0C0}0.098\std0.002 & \multicolumn{1}{c}{\cellcolor[HTML]{C0C0C0}0.082\std0.002} & \cellcolor[HTML]{C0C0C0}0.151\std0.004 \\  
 & Shap & \multicolumn{1}{c}{0.670\std0.012} & \multicolumn{1}{c}{0.200\std0.024} & \multicolumn{1}{c}{0.058\std0.003} & 0.099\std0.005 & \multicolumn{1}{c}{0.087\std0.008} & 0.146\std0.011 \\  
 & \cellcolor[HTML]{C0C0C0}LLM-ITG & \multicolumn{1}{c}{\cellcolor[HTML]{C0C0C0}0.690\std0.004} & \multicolumn{1}{c}{\cellcolor[HTML]{C0C0C0}0.247\std0.008} & \multicolumn{1}{c}{\cellcolor[HTML]{C0C0C0}0.056\std0.001} & \cellcolor[HTML]{C0C0C0}0.099\std0.002 & \multicolumn{1}{c}{\cellcolor[HTML]{C0C0C0}0.085\std0.002} & \cellcolor[HTML]{C0C0C0}0.148\std0.004 \\  
\multirow{-12}{*}{Recidivism} & ITG & \multicolumn{1}{c}{0.689\std0.011} & \multicolumn{1}{c}{0.195\std0.022} & \multicolumn{1}{c}{0.056\std0.003} & 0.100\std0.005 & \multicolumn{1}{c}{0.078\std0.007} & 0.149\std0.011 \\ \hline
 & \cellcolor[HTML]{C0C0C0}LLM-Lime & \multicolumn{1}{c}{\cellcolor[HTML]{C0C0C0}0.909\std0.001} & \multicolumn{1}{c}{\cellcolor[HTML]{C0C0C0}0.632\std0.005} & \multicolumn{1}{c}{\cellcolor[HTML]{C0C0C0}0.023\std0.001} & \cellcolor[HTML]{C0C0C0}0.222\std0.003 & \multicolumn{1}{c}{\cellcolor[HTML]{C0C0C0}0.035\std0.002} & \cellcolor[HTML]{C0C0C0}0.230\std0.004 \\  
 & Lime & \multicolumn{1}{c}{0.907\std0.005} & \multicolumn{1}{c}{0.743\std0.017} & \multicolumn{1}{c}{0.018\std0.002} & 0.224\std0.011 & \multicolumn{1}{c}{0.029\std0.005} & 0.235\std0.014 \\  
 & \cellcolor[HTML]{C0C0C0}LLM-Grad & \multicolumn{1}{c}{\cellcolor[HTML]{C0C0C0}0.938\std0.000} & \multicolumn{1}{c}{\cellcolor[HTML]{C0C0C0}0.801\std0.001} & \multicolumn{1}{c}{\cellcolor[HTML]{C0C0C0}0.022\std0.001} & \cellcolor[HTML]{C0C0C0}0.223\std0.003 & \multicolumn{1}{c}{\cellcolor[HTML]{C0C0C0}0.035\std0.002} & \cellcolor[HTML]{C0C0C0}0.230\std0.004 \\  
 & Grad & \multicolumn{1}{c}{0.999\std0.001} & \multicolumn{1}{c}{0.997\std0.003} & \multicolumn{1}{c}{0.018\std0.002} & 0.224\std0.011 & \multicolumn{1}{c}{0.029\std0.004} & 0.234\std0.014 \\  
 & \cellcolor[HTML]{C0C0C0}LLM-SG & \multicolumn{1}{c}{\cellcolor[HTML]{C0C0C0}0.938\std0.000} & \multicolumn{1}{c}{\cellcolor[HTML]{C0C0C0}0.802\std0.001} & \multicolumn{1}{c}{\cellcolor[HTML]{C0C0C0}0.022\std0.001} & \cellcolor[HTML]{C0C0C0}0.223\std0.003 & \multicolumn{1}{c}{\cellcolor[HTML]{C0C0C0}0.035\std0.002} & \cellcolor[HTML]{C0C0C0}0.230\std0.004 \\  
 & SG & \multicolumn{1}{c}{0.999\std0.001} & \multicolumn{1}{c}{0.997\std0.003} & \multicolumn{1}{c}{0.018\std0.002} & 0.224\std0.011 & \multicolumn{1}{c}{0.029\std0.004} & 0.234\std0.014 \\  
 & \cellcolor[HTML]{C0C0C0}LLM-IG & \multicolumn{1}{c}{\cellcolor[HTML]{C0C0C0}0.938\std0.000} & \multicolumn{1}{c}{\cellcolor[HTML]{C0C0C0}0.804\std0.000} & \multicolumn{1}{c}{\cellcolor[HTML]{C0C0C0}0.022\std0.001} & \cellcolor[HTML]{C0C0C0}0.223\std0.003 & \multicolumn{1}{c}{\cellcolor[HTML]{C0C0C0}0.033\std0.002} & \cellcolor[HTML]{C0C0C0}0.231\std0.004 \\  
 & IG & \multicolumn{1}{c}{1.000\std0.000} & \multicolumn{1}{c}{1.000\std0.000} & \multicolumn{1}{c}{0.018\std0.002} & 0.224\std0.011 & \multicolumn{1}{c}{0.031\std0.005} & 0.235\std0.014 \\  
 & \cellcolor[HTML]{C0C0C0}LLM-Shap & \multicolumn{1}{c}{\cellcolor[HTML]{C0C0C0}0.676\std0.002} & \multicolumn{1}{c}{\cellcolor[HTML]{C0C0C0}0.069\std0.003} & \multicolumn{1}{c}{\cellcolor[HTML]{C0C0C0}0.109\std0.002} & \cellcolor[HTML]{C0C0C0}0.148\std0.003 & \multicolumn{1}{c}{\cellcolor[HTML]{C0C0C0}0.123\std0.003} & \cellcolor[HTML]{C0C0C0}0.153\std0.004 \\  
 & Shap & \multicolumn{1}{c}{0.662\std0.007} & \multicolumn{1}{c}{0.107\std0.012} & \multicolumn{1}{c}{0.139\std0.009} & 0.127\std0.009 & \multicolumn{1}{c}{0.144\std0.011} & 0.149\std0.013 \\  
 & \cellcolor[HTML]{C0C0C0}LLM-ITG & \multicolumn{1}{c}{\cellcolor[HTML]{C0C0C0}0.665\std0.002} & \multicolumn{1}{c}{\cellcolor[HTML]{C0C0C0}0.039\std0.002} & \multicolumn{1}{c}{\cellcolor[HTML]{C0C0C0}0.107\std0.002} & \cellcolor[HTML]{C0C0C0}0.150\std0.003 & \multicolumn{1}{c}{\cellcolor[HTML]{C0C0C0}0.132\std0.003} & \cellcolor[HTML]{C0C0C0}0.146\std0.004 \\  
\multirow{-12}{*}{Adult} & ITG & \multicolumn{1}{c}{0.627\std0.006} & \multicolumn{1}{c}{0.068\std0.010} & \multicolumn{1}{c}{0.175\std0.010} & 0.099\std0.009 & \multicolumn{1}{c}{0.170\std0.011} & 0.130\std0.013 \\ \hline
 & \cellcolor[HTML]{C0C0C0}LLM-Lime & \multicolumn{1}{c}{\cellcolor[HTML]{C0C0C0}0.954\std0.001} & \multicolumn{1}{c}{\cellcolor[HTML]{C0C0C0}0.787\std0.003} & \multicolumn{1}{c}{\cellcolor[HTML]{C0C0C0}0.030\std0.001} & \cellcolor[HTML]{C0C0C0}0.189\std0.003 & \multicolumn{1}{c}{\cellcolor[HTML]{C0C0C0}0.042\std0.002} & \cellcolor[HTML]{C0C0C0}0.178\std0.003 \\  
 & Lime & \multicolumn{1}{c}{0.977\std0.004} & \multicolumn{1}{c}{0.878\std0.015} & \multicolumn{1}{c}{0.030\std0.003} & 0.201\std0.009 & \multicolumn{1}{c}{0.037\std0.004} & 0.186\std0.010 \\  
 & \cellcolor[HTML]{C0C0C0}LLM-Grad & \multicolumn{1}{c}{\cellcolor[HTML]{C0C0C0}0.984\std0.000} & \multicolumn{1}{c}{\cellcolor[HTML]{C0C0C0}0.896\std0.001} & \multicolumn{1}{c}{\cellcolor[HTML]{C0C0C0}0.029\std0.001} & \cellcolor[HTML]{C0C0C0}0.189\std0.003 & \multicolumn{1}{c}{\cellcolor[HTML]{C0C0C0}0.042\std0.002} & \cellcolor[HTML]{C0C0C0}0.178\std0.003 \\  
 & Grad & \multicolumn{1}{c}{1.000\std0.000} & \multicolumn{1}{c}{1.000\std0.000} & \multicolumn{1}{c}{0.030\std0.003} & 0.201\std0.009 & \multicolumn{1}{c}{0.038\std0.005} & 0.185\std0.011 \\  
 & \cellcolor[HTML]{C0C0C0}LLM-SG & \multicolumn{1}{c}{\cellcolor[HTML]{C0C0C0}0.984\std0.000} & \multicolumn{1}{c}{\cellcolor[HTML]{C0C0C0}0.897\std0.000} & \multicolumn{1}{c}{\cellcolor[HTML]{C0C0C0}0.029\std0.001} & \cellcolor[HTML]{C0C0C0}0.189\std0.003 & \multicolumn{1}{c}{\cellcolor[HTML]{C0C0C0}0.072\std0.003} & \cellcolor[HTML]{C0C0C0}0.165\std0.003 \\  
 & SG & \multicolumn{1}{c}{1.000\std0.000} & \multicolumn{1}{c}{1.000\std0.000} & \multicolumn{1}{c}{0.030\std0.003} & 0.201\std0.009 & \multicolumn{1}{c}{0.037\std0.004} & 0.185\std0.011 \\  
 & \cellcolor[HTML]{C0C0C0}LLM-IG & \multicolumn{1}{c}{\cellcolor[HTML]{C0C0C0}0.984\std0.000} & \multicolumn{1}{c}{\cellcolor[HTML]{C0C0C0}0.896\std0.001} & \multicolumn{1}{c}{\cellcolor[HTML]{C0C0C0}0.029\std0.001} & \cellcolor[HTML]{C0C0C0}0.189\std0.003 & \multicolumn{1}{c}{\cellcolor[HTML]{C0C0C0}0.041\std0.002} & \cellcolor[HTML]{C0C0C0}0.179\std0.003 \\  
 & IG & \multicolumn{1}{c}{1.000\std0.000} & \multicolumn{1}{c}{1.000\std0.000} & \multicolumn{1}{c}{0.030\std0.003} & 0.201\std0.009 & \multicolumn{1}{c}{0.041\std0.005} & 0.185\std0.010 \\  
 & \cellcolor[HTML]{C0C0C0}LLM-Shap & \multicolumn{1}{c}{\cellcolor[HTML]{C0C0C0}0.543\std0.003} & \multicolumn{1}{c}{\cellcolor[HTML]{C0C0C0}0.067\std0.004} & \multicolumn{1}{c}{\cellcolor[HTML]{C0C0C0}0.088\std0.002} & \cellcolor[HTML]{C0C0C0}0.140\std0.003 & \multicolumn{1}{c}{\cellcolor[HTML]{C0C0C0}0.094\std0.003} & \cellcolor[HTML]{C0C0C0}0.126\std0.003 \\  
 & Shap & \multicolumn{1}{c}{0.525\std0.009} & \multicolumn{1}{c}{0.086\std0.012} & \multicolumn{1}{c}{0.088\std0.005} & 0.163\std0.010 & \multicolumn{1}{c}{0.091\std0.006} & 0.146\std0.011 \\  
 & \cellcolor[HTML]{C0C0C0}LLM-ITG & \multicolumn{1}{c}{\cellcolor[HTML]{C0C0C0}0.526\std0.003} & \multicolumn{1}{c}{\cellcolor[HTML]{C0C0C0}0.052\std0.003} & \multicolumn{1}{c}{\cellcolor[HTML]{C0C0C0}0.088\std0.002} & \cellcolor[HTML]{C0C0C0}0.139\std0.003 & \multicolumn{1}{c}{\cellcolor[HTML]{C0C0C0}0.091\std0.002} & \cellcolor[HTML]{C0C0C0}0.129\std0.003 \\  
\multirow{-12}{*}{\makecell{\credit}} & ITG & \multicolumn{1}{c}{0.516\std0.010} & \multicolumn{1}{c}{0.076\std0.012} & \multicolumn{1}{c}{0.086\std0.005} & 0.165\std0.010 & \multicolumn{1}{c}{0.084\std0.006} & 0.152\std0.010 \\
\bottomrule
\end{tabular}\label{tab:exp_based_icl}
\end{table*}

\begin{table*}[h]
\centering
\setlength{\tabcolsep}{0pt}
\renewcommand{\arraystretch}{1.1}
\footnotesize
\caption{Average faithfulness metric values (PGI-text and PGU-text) and their standard errors for explanations across 100 test instances. It compares explanations from \vanillaExp and \instructionICL using \gptfour and \gptfournew for the top-$k = 3$ most important words on the Yelp, IMDb, and Amazon datasets. We calculated each metric's area under the curve (AUC) for $k=3$ (where the AUC is calculated from $k=1$ to $k=3$).
Arrows ($\uparrow$,
$\downarrow$) denote the direction of improved performance.}
\begin{tabular}{lcccccc}
\toprule
\cellcolor[HTML]{FFFFFF}                         & \multicolumn{2}{c}{\cellcolor[HTML]{FFFFFF}Yelp}                  & \multicolumn{2}{c}{\cellcolor[HTML]{FFFFFF}IMDb}                 & \multicolumn{2}{c}{\cellcolor[HTML]{FFFFFF}Amazon}               \\
\multirow{-2}{*}{\cellcolor[HTML]{FFFFFF}Method} & \multicolumn{1}{c}{PGI-text ($\uparrow$)} & \multicolumn{1}{c}{PGU-text ($\downarrow$)} & \multicolumn{1}{c}{PGI-text ($\uparrow$)} & \multicolumn{1}{c}{PGU-text ($\downarrow$)} & \multicolumn{1}{c}{PGI-text ($\uparrow$)} & \multicolumn{1}{c}{PGU-text ($\downarrow$)} \\
\midrule
PG-ICL (\gptfournew)                                  & 0.244\std0.035                   & 0.260\std0.034                    & 0.329\std0.045                   & 0.214\std0.039                    & 0.371\std0.042                   & 0.205\std0.035                            \\
P-ICL (\gptfournew)                                   & 0.300\std0.038                   & 0.231\std0.033                             & 0.295\std0.040                   & 0.263\std0.039                    & 0.390\std0.041                   & 0.252\std0.037                   \\
PG-ICL (\gptfour)                                    & 0.349\std0.039                            & 0.243\std0.033                             & 0.318\std0.043                   & 0.251\std0.038                    & 0.386\std0.038                   & 0.238\std0.036                   \\
P-ICL (\gptfour)                                     & 0.300\std0.036                   & 0.293\std0.037                    & 0.332\std0.038                             & 0.205\std0.031                             & 0.409\std0.038                             & 0.222\std0.036                   \\
LA                                                & 0.212\std0.035                   & 0.371\std0.041                    & 0.140\std0.028                   & 0.418\std0.048                    & 0.218\std0.034                   & 0.290\std0.038                   \\
LC                                                & 0.170\std0.030                   & 0.355\std0.042                    & 0.178\std0.033                   & 0.339\std0.042                    & 0.216\std0.035                   & 0.303\std0.037                   \\
LDL                                               & 0.301\std0.038                   & 0.305\std0.038                    & 0.208\std0.034                   & 0.367\std0.046                    & 0.297\std0.040                   & 0.265\std0.038                   \\
LGS                                               & 0.291\std0.039                   & 0.355\std0.038                    & 0.186\std0.033                   & 0.375\std0.046                    & 0.240\std0.035                   & 0.313\std0.040                   \\
LGxA                                              & 0.280\std0.038                   & 0.331\std0.042                    & 0.229\std0.037                   & 0.373\std0.045                    & 0.327\std0.042                   & 0.272\std0.039                   \\
LIG                                               & 0.291\std0.040                   & 0.346\std0.040                    & 0.217\std0.035                   & 0.394\std0.047                    & 0.263\std0.040                   & 0.358\std0.042                   \\
LIME                                              & 0.382\std0.038                             & 0.239\std0.033                             & 0.486\std0.039                             & 0.163\std0.031                             & 0.429\std0.040                             & 0.181\std0.031                             \\
LIME-16                                           & 0.380\std0.041                             & 0.258\std0.033                    & 0.359\std0.044                             & 0.201\std0.034                             & 0.396\std0.041                             & 0.171\std0.032                             \\
Random                                            & 0.271\std0.029                   & 0.251\std0.027                    & 0.291\std0.032                   & 0.284\std0.031                    & 0.304\std0.032                   & 0.281\std0.034                   \\
\bottomrule
\end{tabular}
\label{tab:new_scores}
\end{table*}

\begin{table*}[h]
\centering
\setlength{\tabcolsep}{2.1pt}
\renewcommand{\arraystretch}{1.1}
\footnotesize
\caption{Faithfulness scores for the most important feature value, top-$k=1$, identified by existing post hoc explanation methods as well as the three LLM methods which generated explanations from \gptfour across four datasets and the LR model. (Since FA = RA for top-$k=1$, we omit RA to avoid redundancy).
}
\begin{tabular}{lcccccccc}
\toprule
\cellcolor[HTML]{FFFFFF}                         & \multicolumn{2}{c}{\cellcolor[HTML]{FFFFFF}Recidivism}                    & \multicolumn{2}{c}{Adult}                                                 & \multicolumn{2}{c}{Credit}                                                & \multicolumn{2}{c}{Blood}                                                 \\
\multirow{-2}{*}{\cellcolor[HTML]{FFFFFF}Method} & \multicolumn{1}{c}{FA ($\uparrow$)} & \multicolumn{1}{c}{PGU ($\downarrow$)} & \multicolumn{1}{c}{FA ($\uparrow$)} & \multicolumn{1}{c}{PGU ($\downarrow$)}& \multicolumn{1}{c}{FA ($\uparrow$)} & \multicolumn{1}{c}{PGU ($\downarrow$)} & \multicolumn{1}{c}{FA ($\uparrow$)} & \multicolumn{1}{c}{PGU ($\downarrow$)} \\
\midrule
Grad                                             & 1.000\std0.000                           & 0.096\std0.005                           & 1.000\std0.000                           & 0.073\std0.007                           & 1.000\std0.000                           & 0.081\std0.006                           & 1.000\std0.000                           & 0.020\std0.000                           \\
SG                                               & 1.000\std0.000                           & 0.095\std0.005                           & 1.000\std0.000                           & 0.073\std0.007                           & 1.000\std0.000                           & 0.081\std0.006                           & 1.000\std0.000                           & 0.020\std0.000                           \\
IG                                               & 1.000\std0.000                           & 0.096\std0.005                           & 1.000\std0.000                           & 0.073\std0.007                           & 1.000\std0.000                           & 0.081\std0.006                           & 1.000\std0.000                           & 0.020\std0.000                           \\
ITG                                              & 0.190\std0.039                        & 0.108\std0.006                        & 0.020\std0.014                        & 0.221\std0.011                        & 0.270\std0.044                        & 0.163\std0.007                        & 0.700\std0.046                         & 0.026\std0.001                         \\
SHAP                                             & 0.210\std0.041                        & 0.108\std0.006                        & 0.020\std0.014                        & 0.221\std0.011                        & 0.270\std0.044                        & 0.163\std0.007                        & 0.700\std0.046                         & 0.026\std0.001                        \\
LIME                                             & 1.000\std0.000                           & 0.096\std0.005                           & 0.990\std0.010                         & 0.221\std0.011                         & 1.000\std0.000                           & 0.081\std0.006                           & 1.000\std0.000                           & 0.020\std0.000                          \\
Random                                           & 0.130\std0.034                        & 0.113\std0.006                        & 0.060\std0.024                        & 0.214\std0.011                        & 0.070\std0.026                        & 0.195\std0.010                        & 0.190\std0.039                        & 0.038\std0.001    \\
P-ICL                                  & 0.011\std0.011                          & 0.102\std0.005                         & 0.716\std0.050                          & 0.116\std0.012                       & 1.000\std0.000                         & 0.081\std0.007                       & 
0.988\std0.012                         & 0.020\std0.000                        \\
PG-ICL                                  & 0.269\std0.046                          & 0.101\std0.005                        & 0.869\std0.034                          & 0.094\std0.009                       & 
0.918\std0.028                         & 0.092\std0.008                       & 0.845\std0.039                         & 0.023\std0.001                        \\
E-ICL                                  & 0.490\std0.050                          & 0.098\std0.005                        & 0.919\std0.027                          & 0.086\std0.009                       & 0.926\std0.027                         & 0.090\std0.007                       & 0.758\std0.045                         & 0.025\std0.001                        \\
\bottomrule
\end{tabular}\label{tab:top1_lr}
\end{table*}



\begin{table*}[h]
\centering
\setlength{\tabcolsep}{2.1pt}
\renewcommand{\arraystretch}{1.1}
\footnotesize
\caption{Faithfulness scores for the most important feature value, top-$k=1$, identified by existing post hoc explanation methods as well as the three LLM methods which generated explanations from \gptfour across four datasets and the ANN model.
}
\begin{tabular}{lcccccccc}
\toprule
\cellcolor[HTML]{FFFFFF}                         & \multicolumn{2}{c}{\cellcolor[HTML]{FFFFFF}Recidivism}                        & \multicolumn{2}{c}{Adult}                                                     & \multicolumn{2}{c}{Credit}                                                    & \multicolumn{2}{c}{Blood}                                                     \\
\multirow{-2}{*}{\cellcolor[HTML]{FFFFFF}Method} & \multicolumn{1}{c}{PGU ($\downarrow$)} & \multicolumn{1}{c}{PGI ($\uparrow$)} & \multicolumn{1}{c}{PGU ($\downarrow$)} & \multicolumn{1}{c}{PGI ($\uparrow$)} & \multicolumn{1}{c}{PGU ($\downarrow$)} & \multicolumn{1}{c}{PGI ($\uparrow$)} & \multicolumn{1}{c}{PGU ($\downarrow$)} & \multicolumn{1}{c}{PGI ($\uparrow$)} \\
\midrule
Grad                                             & 0.147\std0.011                          & 0.117\std0.010                         & 0.103\std0.013                          & 0.224\std0.014                        & 0.085\std0.009                          & 0.166\std0.010                         & 0.087\std0.012                          & 0.103\std0.012                        \\
SG                                               & 0.146\std0.011                          & 0.117\std0.010                         & 0.103\std0.013                          & 0.224\std0.014                        & 0.084\std0.009                          & 0.167\std0.010                         & 0.087\std0.012                          & 0.102\std0.012                        \\
IG                                               & 0.147\std0.011                          & 0.116\std0.010                         & 0.103\std0.013                          & 0.225\std0.014                        & 0.085\std0.009                          & 0.167\std0.010                         & 0.087\std0.012                          & 0.103\std0.012                        \\
ITG                                              & 0.154\std0.012                          & 0.084\std0.009                        & 0.232\std0.014                          & 0.056\std0.009                        & 0.181\std0.010                           & 0.057\std0.009                        & 0.103\std0.012                          & 0.083\std0.012                        \\
SHAP                                             & 0.152\std0.012                          & 0.092\std0.009                        & 0.231\std0.014                          & 0.047\std0.008                        & 0.169\std0.009                          & 0.076\std0.011                        & 0.104\std0.012                          & 0.083\std0.012                        \\
LIME                                             & 0.147\std0.011                          & 0.116\std0.010                         & 0.104\std0.013                          & 0.225\std0.014                        & 0.084\std0.009                          & 0.167\std0.010                         & 0.087\std0.012                          & 0.103\std0.012                        \\
Random                                           & 0.163\std0.012                          & 0.062\std0.009                        & 0.228\std0.014                          & 0.031\std0.008                        & 0.187\std0.011                          & 0.033\std0.006                        & 0.115\std0.012                          & 0.067\std0.010                        \\
P-ICL & 0.146\std0.012                          & 0.114\std0.010                        & 0.146\std0.016                          & 0.160\std0.015                       & 0.086\std0.009                         & 0.166\std0.011                       & 0.081\std0.012                         & 0.095\std0.013                        \\
PG-ICL  & 0.146\std0.012                          & 0.113\std0.010                        & 0.138\std0.015                          & 0.179\std0.015                       & 0.095\std0.010                         & 0.155\std0.011                       & 0.086\std0.013                         & 0.089\std0.013                        \\
E-ICL  & 0.148\std0.011                          & 0.114\std0.010                        & 0.103\std0.013                          & 0.224\std0.014                       & 0.085\std0.009                         & 0.167\std0.010                       & 0.089\std0.012                         & 0.101\std0.012                        \\
\bottomrule
\end{tabular}\label{tab:top1_ann}
\end{table*}

\newpage
\subsection{LLM Replies}
\label{app:replies}
We provide five example prompts below on the \blood, \adult, and IMDb datasets for \pc. 
See Fig.~\ref{fig:bloodcorrect} for a correct reply and Fig.~\ref{fig:bloodincorrect} for an incorrect reply on the \blood dataset. 
See Fig.~\ref{fig:adultcorrect} for a partially correct reply and Fig.~\ref{fig:adultincorrect} for an incorrect reply on the \adult dataset. See Fig.~\ref{fig:sentiment_classification} for an example reply on the IMDb sentiment classification task.

\begin{figure*}[h]
\begin{ourbox}
\noindent\footnotesize\textbf{A correct sample LLM reply for the Blood dataset:}

\footnotesize Context: ``We have a two-class machine learning model that predicts based on 4 features: [`A', `B', `C', `D']. The dataset below contains the change in feature values `A' through `D' with respect to a given instance and the corresponding change in model outputs.''

\tiny\noindent Dataset:
\begin{verbatim}
Change in Input: A: 0.260, B: -0.037, C: -0.287, D: 0.040
Change in Output: 0

Change in Input: A: 0.283, B: 0.067, C: -0.186, D: -0.099
Change in Output: 0

Change in Input: A: -0.302, B: -0.051, C: 0.232, D: 0.156
Change in Output: 1

Change in Input: A: -0.273, B: 0.082, C: 0.176, D: -0.065
Change in Output: 1

Change in Input: A: -0.210, B: 0.059, C: 0.272, D: 0.045
Change in Output: 1

Change in Input: A: 0.311, B: -0.081, C: -0.111, D: -0.072
Change in Output: 0

Change in Input: A: 0.337, B: -0.010, C: -0.110, D: -0.035
Change in Output: 0

Change in Input: A: -0.326, B: 0.092, C: 0.125, D: 0.082
Change in Output: 1

Change in Input: A: -0.285, B: 0.204, C: 0.085, D: -0.099
Change in Output: 1

Change in Input: A: 0.373, B: 0.173, C: -0.074, D: 0.014
Change in Output: 0

Change in Input: A: -0.266, B: 0.173, C: 0.142, D: 0.106
Change in Output: 1

Change in Input: A: -0.382, B: 0.211, C: 0.001, D: -0.034
Change in Output: 1

Change in Input: A: 0.340, B: -0.057, C: -0.048, D: 0.174
Change in Output: 0

Change in Input: A: 0.270, B: -0.083, C: -0.148, D: -0.086
Change in Output: 0

Change in Input: A: 0.203, B: -0.100, C: -0.434, D: -0.127
Change in Output: 0

Change in Input: A: -0.227, B: -0.049, C: 0.322, D: 0.011
Change in Output: 1
\end{verbatim}

\footnotesize\noindent Question: ``Based on the above set, what are the four most important features driving the output?"

\medskip

\footnotesize\noindent Instructions: ``Think about the question. After explaining your reasoning, provide your answer as the top four features ranked from most important to least important, in descending order, separated by commas. Only provide the feature names on the last line. Do not provide any further details on the last line."

\medskip

\footnotesize\noindent\textbf{REPLY:}\\
The question asks for the four most important features driving the output. In this case, we only have four features: `A', `B', `C', and `D'. Therefore, all of these features are important in driving the output. 

However, to rank them in order of importance, we need to look at the changes in input values and how they correspond to the changes in output. 

From the dataset, we can observe that:

- When `A' increases, the output tends to be 0. When `A' decreases, the output tends to be 1.
- `B' seems to have less clear correlation with the output, as it increases and decreases in both output cases.
- When `C' decreases, the output tends to be 0. When `C' increases, the output tends to be 1.
- `D' also seems to have less clear correlation with the output, as it increases and decreases in both output cases.

Based on these observations, it seems that `A' and `C' have the strongest correlation with the output, while `B' and `D' have less clear correlations. Therefore, we can rank the features in the following order of importance: `A', `C', `B', `D'.

\noindent A, C, B, D
\end{ourbox}
\caption{A correct sample LLM reply for the Blood dataset. The ground-truth feature ranking is A, C, B, D.}
\label{fig:bloodcorrect}
\end{figure*}

\begin{figure*}[h]
\begin{ourbox}
\noindent\footnotesize\textbf{An incorrect sample LLM reply for the Blood dataset}

\footnotesize Context: ``We have a two-class machine learning model that predicts based on 4 features: [`A', `B', `C', `D']. The dataset below contains the change in feature values `A' through `D' with respect to a given instance and the corresponding change in model outputs.''

\tiny\noindent Dataset:
\begin{verbatim}
Change in Input: A: 0.380, B: -0.185, C: -0.026, D: 0.065
Change in Output: -1

Change in Input: A: 0.277, B: -0.144, C: -0.157, D: -0.061
Change in Output: -1

Change in Input: A: -0.395, B: 0.114, C: 0.045, D: -0.104
Change in Output: 0

Change in Input: A: -0.406, B: 0.098, C: -0.099, D: -0.044
Change in Output: 0

Change in Input: A: -0.251, B: 0.023, C: 0.241, D: 0.015
Change in Output: 0

Change in Input: A: 0.331, B: -0.052, C: -0.165, D: -0.002
Change in Output: -1

Change in Input: A: 0.341, B: -0.003, C: -0.160, D: -0.081
Change in Output: -1

Change in Input: A: -0.242, B: 0.085, C: 0.302, D: 0.118
Change in Output: 0

Change in Input: A: -0.347, B: 0.110, C: 0.036, D: -0.071
Change in Output: 0

Change in Input: A: 0.290, B: -0.114, C: -0.143, D: -0.078
Change in Output: -1

Change in Input: A: -0.294, B: 0.061, C: 0.154, D: 0.021
Change in Output: 0
\end{verbatim}

\footnotesize\noindent Question: ``Based on the above set, what are the four most important features driving the output?"

\medskip

\footnotesize\noindent Instructions: ``Think about the question. After explaining your reasoning, provide your answer as the top four features ranked from most important to least important, in descending order, separated by commas. Only provide the feature names on the last line. Do not provide any further details on the last line."

\medskip

\footnotesize\noindent\textbf{REPLY:}\\The question asks for the four most important features driving the output. In this case, we only have four features: `A', `B', `C', and `D'. Therefore, all of them are important. However, to rank them, we need to look at how changes in these features affect the output.

From the dataset, we can observe that:

1. Feature `A': When `A' increases, the output tends to decrease (output is -1). When `A' decreases, the output tends to stay the same (output is 0). This suggests that `A' has a strong influence on the output.

2. Feature `B': When `B' decreases, the output tends to decrease (output is -1). When `B' increases, the output tends to stay the same (output is 0). This suggests that `B' also has a strong influence on the output.

3. Feature `C': When `C' decreases, the output tends to decrease (output is -1). When `C' increases, the output tends to stay the same (output is 0). This suggests that `C' also has a strong influence on the output.

4. Feature `D': The changes in `D' do not seem to have a consistent effect on the output. Sometimes when `D' increases or decreases, the output stays the same (output is 0), and other times it decreases (output is -1). This suggests that `D' has a weaker influence on the output compared to `A', `B', and `C'.

Therefore, based on the given dataset, the four features can be ranked in importance as follows:

A, B, C, D

\noindent
\end{ourbox}
\caption{An incorrect sample LLM reply for the \blood dataset. The ground-truth feature ranking is A, C, B, D.}
\label{fig:bloodincorrect}
\end{figure*}

\begin{figure*}[h]
\begin{ourbox}
\noindent\footnotesize\textbf{A partially correct sample LLM reply for the \adult dataset:}

\footnotesize Context: "We have a two-class machine learning model that predicts based on 13 features: [`A', `B', `C', `D', `E', `F', `G', `H', `I', `J', `K', `L', `M']. The dataset below contains the change in feature values `A' through `M' with respect to a given instance and the corresponding change in model outputs."

\tiny\noindent Dataset:
\begin{verbatim}
Change in Input: A: -0.015, B: 0.014, C: 0.100, D: -0.382, E: -0.142, F: -0.063, G: 0.143, H: 0.005, I: 0.022, J: -0.150, 
                 K: 0.054, L: 0.142, M: 0.099
Change in Output: 0

Change in Input: A: -0.073, B: -0.006, C: 0.042, D: -0.336, E: -0.006, F: -0.081, G: -0.092, H: -0.089, I: 0.053, J: 0.216, 
                 K: -0.003, L: -0.118, M: -0.007
Change in Output: 0

Change in Input: A: -0.052, B: 0.131, C: 0.007, D: 0.387, E: 0.064, F: 0.126, G: -0.026, H: -0.118, I: 0.025, J: 0.080, 
                 K: 0.070, L: 0.052, M: 0.018
Change in Output: 1

Change in Input: A: 0.060, B: -0.023, C: 0.220, D: 0.235, E: 0.063, F: 0.160, G: -0.113, H: 0.267, I: -0.190, J: 0.010, 
                 K: -0.071, L: 0.063, M: -0.092
Change in Output: 1

Change in Input: A: -0.043, B: -0.066, C: 0.055, D: 0.306, E: 0.076, F: 0.055, G: -0.048, H: -0.144, I: 0.001, J: -0.064, 
                 K: -0.157, L: 0.108, M: 0.210
Change in Output: 1

Change in Input: A: -0.061, B: 0.174, C: 0.010, D: -0.374, E: 0.056, F: 0.074, G: -0.076, H: 0.134, I: 0.097, J: -0.096, 
                 K: -0.147, L: 0.013, M: 0.179
Change in Output: 0

Change in Input: A: -0.201, B: -0.031, C: -0.184, D: -0.319, E: -0.096, F: -0.153, G: 0.030, H: 0.124, I: -0.030, J: -0.130,  
                 K: -0.011, L: -0.059, M: -0.117
Change in Output: 0

Change in Input: A: 0.070, B: -0.032, C: 0.103, D: 0.326, E: 0.245, F: 0.096, G: 0.119, H: 0.125, I: 0.021, J: -0.038,  
                 K: 0.107, L: 0.101, M: 0.273
Change in Output: 1

Change in Input: A: 0.039, B: 0.074, C: -0.020, D: 0.337, E: -0.112, F: -0.004, G: -0.019, H: 0.086, I: -0.123, J: 0.086,  
                 K: -0.047, L: -0.287, M: -0.054
Change in Output: 1

Change in Input: A: 0.116, B: -0.083, C: 0.077, D: -0.392, E: 0.071, F: -0.043, G: 0.057, H: 0.076, I: 0.003, J: 0.169,  
                 K: -0.144, L: 0.127, M: -0.074
Change in Output: 0

Change in Input: A: 0.133, B: -0.074, C: 0.195, D: 0.260, E: 0.129, F: 0.003, G: -0.016, H: 0.112, I: -0.156, J: 0.019,  
                 K: -0.031, L: 0.010, M: -0.139
Change in Output: 1

...

Change in Input: A: -0.008, B: 0.099, C: 0.074, D: 0.302, E: 0.228, F: -0.027, G: -0.077, H: 0.005, I: 0.032, J: 0.034,  
                 K: -0.009, L: 0.033, M: -0.131
Change in Output: 1
\end{verbatim}

\footnotesize\noindent Question: ``Based on the above set, what are the four most important features driving the output?"

\medskip

\footnotesize\noindent Instructions: ``Think about the question. After explaining your reasoning, provide your answer as the top four features ranked from most important to least important, in descending order, separated by commas. Only provide the feature names on the last line. Do not provide any further details on the last line."

\medskip

\footnotesize\noindent\textbf{REPLY:}\\
To determine the most important features driving the output, we need to look at the changes in input features and how they correspond to the changes in output. The most important features would be those that, when they change, the output also changes. 

However, it's important to note that machine learning models, especially complex ones, may not have a direct one-to-one correspondence between a single feature change and the output change. The output could be a result of complex interactions between multiple features. 

In this case, we can make a rough estimation by looking at the features that show significant changes when the output changes. 

Upon examining the dataset, it seems that features `D', `C', `A', `E', and `F' often have larger changes compared to other features when the output changes. 

However, this is a rough estimation and may not be entirely accurate. A more accurate method would be to use techniques like permutation importance, SHAP, or partial dependence plots to determine feature importance, which take into account the complex interactions between features.

D, C, A, E, F

\end{ourbox}
\caption{A partially correct sample LLM reply for the \adult dataset. The ground-truth feature ranking is D, C, I, E, F, A, J, B, G, M, L, H, K. Note: the dataset was abbreviated due to space restrictions.}
\label{fig:adultcorrect}
\end{figure*}

\begin{figure*}[h]
\begin{ourbox}\label{app:adult-pc-incorrect}
\noindent\footnotesize\textbf{An incorrect sample LLM reply for the \adult dataset}

\footnotesize Context: "We have a two-class machine learning model that predicts based on 13 features: [`A', `B', `C', `D', `E', `F', `G', `H', `I', `J', `K', `L', `M']. The dataset below contains the change in feature values `A' through `M' with respect to a given instance and the corresponding change in model outputs."

\tiny\noindent Dataset:
\begin{verbatim}
Change in Input: A: -0.163, B: 0.011, C: -0.023, D: -0.285, E: -0.094, F: -0.312, G: 0.063, H: 0.105, I: 0.089, J: -0.004,  
                 K: 0.035, L: 0.141, M: -0.031
Change in Output: 0

Change in Input: A: -0.191, B: 0.008, C: -0.011, D: -0.349, E: 0.085, F: 0.011, G: -0.357, H: -0.107, I: -0.114, J: 0.098,  
                 K: 0.179, L: -0.101, M: 0.104
Change in Output: 0

Change in Input: A: 0.072, B: 0.030, C: 0.023, D: 0.369, E: 0.099, F: 0.099, G: -0.073, H: -0.006, I: 0.022, J: 0.181,  
                 K: 0.033, L: 0.059, M: -0.083
Change in Output: 1

Change in Input: A: 0.031, B: 0.014, C: -0.044, D: 0.304, E: 0.012, F: 0.180, G: 0.061, H: 0.072, I: -0.108, J: 0.015,  
                 K: 0.128, L: 0.135, M: 0.105
Change in Output: 1

Change in Input: A: -0.021, B: -0.066, C: -0.025, D: 0.305, E: 0.169, F: -0.165, G: 0.008, H: -0.063, I: -0.154, J: -0.010,  
                 K: 0.172, L: 0.016, M: -0.104
Change in Output: 1

Change in Input: A: -0.182, B: 0.136, C: 0.015, D: -0.356, E: 0.032, F: 0.004, G: 0.133, H: -0.053, I: -0.110, J: -0.095,  
                 K: 0.131, L: -0.030, M: 0.067
Change in Output: 0

Change in Input: A: -0.170, B: 0.075, C: -0.062, D: 0.312, E: 0.107, F: 0.005, G: -0.199, H: -0.150, I: -0.295, J: -0.037,  
                 K: 0.017, L: -0.084, M: -0.011
Change in Output: 1

Change in Input: A: -0.167, B: 0.123, C: -0.097, D: -0.280, E: -0.064, F: -0.054, G: 0.085, H: -0.057, I: 0.134, J: 0.069,  
                 K: 0.055, L: 0.080, M: -0.050
Change in Output: 0

Change in Input: A: -0.140, B: 0.063, C: -0.092, D: -0.270, E: -0.022, F: -0.117, G: 0.291, H: -0.064, I: 0.252, J: 0.033,  
                 K: -0.107, L: -0.111, M: 0.198
Change in Output: 0

Change in Input: A: -0.013, B: -0.017, C: -0.138, D: -0.399, E: 0.107, F: -0.112, G: 0.137, H: -0.087, I: -0.029, J: -0.040,  
                 K: -0.054, L: -0.042, M: 0.089
Change in Output: 0

...

Change in Input: A: 0.116, B: 0.024, C: -0.109, D: 0.331, E: 0.035, F: -0.004, G: 0.246, H: 0.064, I: -0.052, J: -0.088,  
                 K: -0.029, L: -0.035, M: -0.014
Change in Output: 1
\end{verbatim}

\footnotesize\noindent Question: ``Based on the above set, what are the four most important features driving the output?"

\medskip

\footnotesize\noindent Instructions: ``Think about the question. After explaining your reasoning, provide your answer as the top four features ranked from most important to least important, in descending order, separated by commas. Only provide the feature names on the last line. Do not provide any further details on the last line."

\medskip

\footnotesize\noindent\textbf{REPLY:}\\To determine the most important features driving the output, we need to look at the changes in input features and how they correspond to the changes in output. The most important features would be those that, when they change, the output also changes significantly. 

However, from the given dataset, it's not straightforward to determine the most important features just by looking at the changes in input and output. This is because the output is binary (0 or 1) and doesn't provide a clear measure of how much each feature contributes to the change in output. 

Moreover, the dataset doesn't provide enough information to determine the correlation between each feature and the output. We would need to perform a statistical analysis, such as a correlation analysis or a feature importance analysis using a machine learning algorithm, to accurately determine the most important features.

Without such analysis, any conclusion drawn from this dataset would be purely speculative. For example, one might be tempted to say that features with larger changes are more important, but this would be a hasty conclusion without further analysis.

Therefore, based on the given dataset, it's not possible to accurately determine the five most important features driving the output. 

However, if we were to make an educated guess based on the magnitude of the changes in the features, we might choose `D', `F', `G', `J', and `K'. But again, this is purely speculative and not based on a rigorous analysis. 

D, F, G, J, K

\noindent
\end{ourbox}
\caption{An incorrect sample LLM reply for the \adult dataset. The ground-truth feature ranking is D, C, I, E, F, A, J, B, G, M, L, H, K. Note: the dataset was abbreviated due to space restrictions.}
\label{fig:adultincorrect}
\end{figure*}


\begin{figure*}[h]
\begin{ourbox}\label{app:adult-pc-incorrect}
\noindent\footnotesize\textbf{A sample LLM reply for the IMDb dataset (sentiment classification task)}

\footnotesize Context: ``We are analyzing a fixed set of word removals on a specific sentence to understand the influence on the model’s output. The dataset below contains the words removed from the original sentence and the corresponding change in output."

\tiny\noindent Dataset:
\begin{verbatim}
Original sentence: It looked like a wonderful story.  

Removed words: it looked like wonderful .
Change in output: 1

Removed words: it .
Change in output: 0

Removed words: looked
Change in output: 1

Removed words: story .
Change in output: 0

Removed words: it looked like a wonderful
Change in output: 1

Removed words: looked story
Change in output: 0

Removed words: it like a wonderful story .
Change in output: 1

Removed words: like
Change in output: 0

Removed words: looked a wonderful story
Change in output: 1

Removed words: it looked like story .
Change in output: 0

Removed words: it looked like a wonderful .
Change in output: 1

Removed words: .
Change in output: 0

Removed words: it a wonderful story .
Change in output: 1

Removed words: .
Change in output: 0

Removed words: it looked a
Change in output: 1

Removed words: it
Change in output: 0
\end{verbatim}

\footnotesize\noindent Question: ``Based on the above set, what are the top three most important words driving the output?"

\medskip

\footnotesize\noindent Instructions: ``Think about the question. After explaining your reasoning, provide your answer as the top three most important words ranked from most important to least important, in descending order. Only provide the important words on the last line. Do not provide any further details on the last line. Provide the answer on one line with each word separated by commas."



\medskip

\footnotesize\noindent\textbf{REPLY:}\\The most important words in determining the model's prediction can be identified by looking at the changes in output when these words are removed. If the removal of a word causes a change in output, it suggests that the word is important for the model's prediction. 

In the given dataset, the words 'it', 'looked', and 'like' cause a change in output when removed. This suggests that these words are important for the model's prediction. 

However, it's important to note that the word 'it' is removed twice and both times it causes a change in output. This suggests that 'it' might be the most important word for the model's prediction. 

Therefore, based on the model’s predictions and the given dataset, the top three most important words in determining the model’s prediction, ranked from most important to least important, are 'it', 'looked', and 'like'.

it, looked, like

\noindent
\end{ourbox}
\caption{A sample LLM reply for the IMDb dataset (sentiment classification task).}
\label{fig:sentiment_classification}
\end{figure*}

\newpage

\end{document}